%% file: main.tex
\documentclass{article} 
\usepackage{collas2022_conference,times}

\input{math_commands.tex}

\usepackage{hyperref}
\hypersetup{
    colorlinks=true,
    linkcolor=red,
    filecolor=magenta,      
    urlcolor=blue,
    citecolor=purple,
    pdftitle={Lifelong Robotic Reinforcement Learning by Retaining Experiences},
    pdfpagemode=FullScreen,
    }

\usepackage{multicol}
\usepackage{multirow}
\usepackage{booktabs}
\usepackage{amsmath}
\usepackage{amsfonts}
\usepackage{mathtools}
\usepackage{bbm}
\usepackage{tabularx}
\usepackage{dsfont}
\usepackage{enumitem}

\usepackage{cuted}
\usepackage{caption}
\usepackage{subfig}
\usepackage{wrapfig,booktabs}
\usepackage{array}

\usepackage{algorithm}
\usepackage[noend]{algpseudocode}
\usepackage{varwidth}

\usepackage{graphicx}
\usepackage{xcolor}
\usepackage{todonotes}
\usepackage{wrapfig}

\title{Lifelong Robotic Reinforcement Learning by Retaining Experiences}

\author{Annie Xie and Chelsea Finn\\
Stanford University\\
\texttt{\{anniexie,cbfinn\}@stanford.edu} \\
}

\collasfinalcopy 

\begin{document}

\maketitle

\begin{abstract}
Multi-task learning ideally allows embodied agents such as robots to acquire a diverse repertoire of useful skills. However, many multi-task reinforcement learning efforts assume the agent can collect data from \emph{all} tasks at \emph{all} times, which can be unrealistic for physical agents that can only attend to one task at a time. Motivated by the practical constraints of physical learning systems, this work studies lifelong learning as a more natural multi-task learning setup. We present an approach that effectively leverages data collected from previous tasks to cumulatively and efficiently grow the robot's skill-set. In a series of simulated robotic manipulation experiments, our approach requires less than half the samples than learning each task from scratch, while avoiding the impractical round-robin data collection scheme. On a Franka Emika Panda robot arm, our approach incrementally solves ten challenging tasks, including bottle capping and block insertion.
\end{abstract}

\input{introduction}

\input{related_work}

\input{preliminaries}

\input{method}

\input{experiments}

\input{conclusion}





\bibliography{collas2022_conference}
\bibliographystyle{collas2022_conference}

\appendix
\input{appendix}

\end{document}

%% file: math_commands.tex
\usepackage{amsmath,amsfonts,bm}

\def\eqref#1{equation~\ref{#1}}

\def\1{\bm{1}}

\def\rb{{\textnormal{b}}}

\DeclareMathAlphabet{\mathsfit}{\encodingdefault}{\sfdefault}{m}{sl}
\SetMathAlphabet{\mathsfit}{bold}{\encodingdefault}{\sfdefault}{bx}{n}



%% file: introduction.tex
\begin{figure*}[b]
    \centering
    \noindent
    \includegraphics[width=\textwidth]{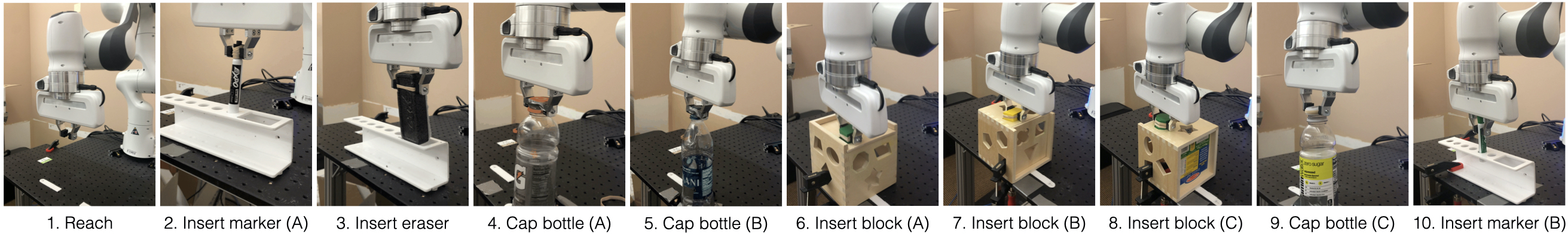}
    \caption{
    \small A Franka Emika robot arm learns a sequence of manipulation tasks, including block insertion and bottle capping, by retaining experience from previous tasks. Our algorithm can learn a sequence of tasks with different setups and objectives with fewer samples compared to when each task is learned from scratch.
    \label{fig:robot_tasks}}
\end{figure*}

\section{Introduction}
\label{sec:introduction}

General-purpose embodied agents ideally should be capable of learning and performing a multitude of tasks. Multi-task learning paves a promising path towards such agents by capitalizing on structure shared between tasks to learn them more efficiently~\citep{rusu2015policy,parisotto2016actor,teh2017distral,hausman2018learning,riedmiller2018learning,yu2020gradient}. However, the standard multi-task reinforcement learning paradigm, which requires data collection from each task in round-robin fashion, can often be unrealistic for embodied agents such as physical robots, which can only attend to one physical workspace at a time. Beyond the impracticality of the framework, it also limits the flexibility in the way tasks are assigned. A robot may encounter new tasks that it needs to learn over the course of its lifetime. For example, the demands of factory robots can vary based on the product being built; the capabilities of service robots may need to grow to accommodate a larger client base. Instead, a more natural multi-task learning setup for physical agents is that of lifelong learning, wherein tasks are learned \emph{in sequence}. 

Two core challenges of lifelong learning are to enable forward transfer, i.e. reusing knowledge from previous tasks to more efficiently learn new tasks, while mitigating negative backward transfer or catastrophic forgetting~\citep{mccloskey1989catastrophic,kirkpatrick2017overcoming}. Prior works have shown that the latter can be addressed by maintaining and replaying prior experiences~\citep{isele2018selective,rolnick2018experience,chaudhry2019tiny,buzzega2020dark,balaji2020effectiveness}. Since real robotic systems are more often bottlenecked by the the time it takes to collect data than by hard drive memory, we adopt this replay approach, assuming complete memory of past experiences, and focus on former challenge of enabling forward transfer across a sequence of tasks.

Prior work that address the forward transfer aspect of the lifelong learning problem primarily focus on the transfer of weights~\citep{caruana1997multitask,fernandez2006probabilistic,rusu2016progressive,schwarz2018progress,julian2020efficient}, e.g., by fine-tuning the previously trained neural network weights that represent the value function or policy. However, in retaining only the neural network weights, these methods may discard important information from the past experiences that are relevant to future tasks. Thus, our work aims to additionally transfer the previously collected low-level experience, in the form of $(\mathbf{s}, \mathbf{a}, \mathbf{s}', r)$ tuples saved in the replay buffer. Rather than leveraging the past experiences with standard experience replay, we propose to prioritize samples based on their utility for new tasks.

Since each task can significantly differ in their transition dynamics and reward functions, naively training on the previously collected samples suffers from sample selection bias~\citep{cortes2008sample}. Hence, when drawing upon samples from previous tasks, we should correct this bias by selectively identifying the most relevant samples for the new task. We measure the similarity between the past samples and the current task's transition dynamics to determine which samples to transfer in the online fine-tuning phase. Consequently, previously-collected samples that are relevant to the target task are used to simulate sampling in the real environment, both accelerating and stabilizing fine-tuning. This method of experience transfer, as we show in our experiments, complements and can be combined with other mediums of transfer such as of the previously trained weights of policy networks.

\begin{figure*}
    \centering
    \vspace{0.6cm}
    \includegraphics[width=0.9\linewidth]{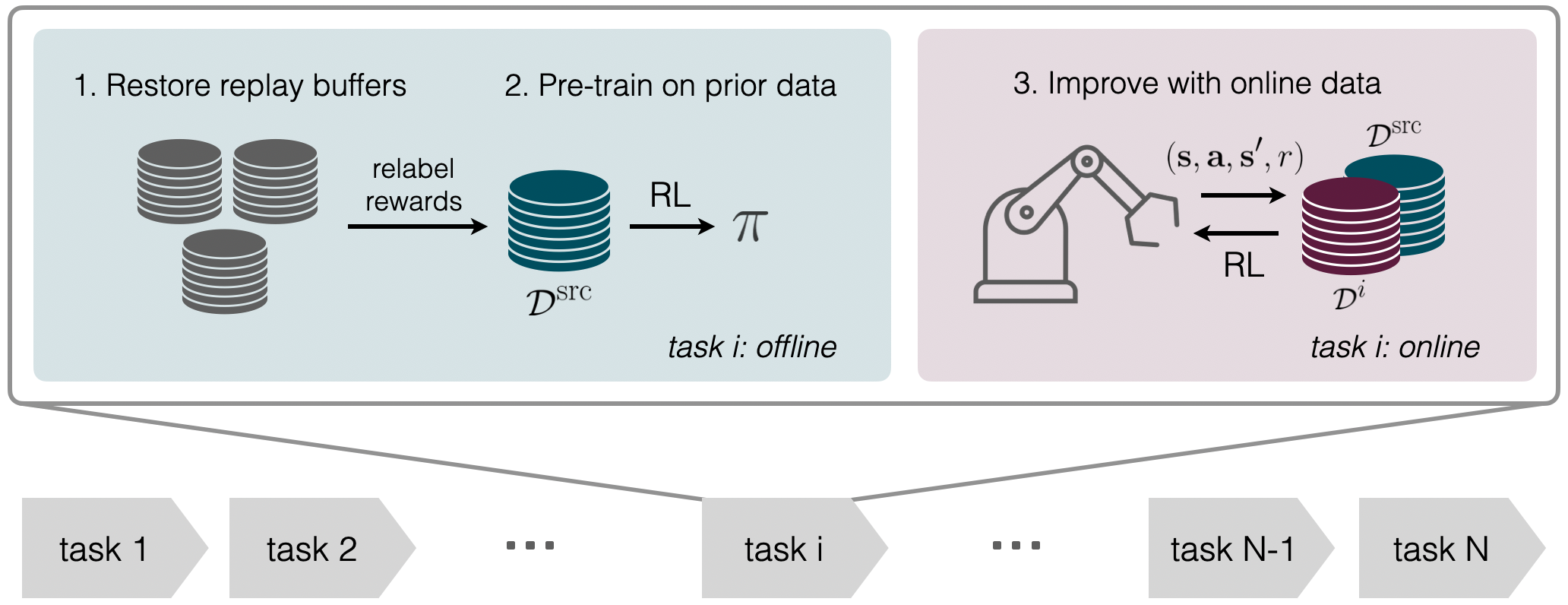}
    \caption{\small In our framework, we perform two steps during each new task. First, we pre-train on the experience from earlier tasks (left). To align the data to the same objective, we use the underlying reward function of the upcoming task to relabel this experience before pre-training. Second, we learn online in the robot's physical environment and gather new data to continuously improve the robot's policy until the task is solved (right).}
    \label{fig:framework}
    \vspace{-0.3cm}
\end{figure*}

The core contribution of this work is a framework for efficient lifelong reinforcement learning that is suitable for physical robots (depicted in Fig.~\ref{fig:framework}). In particular, our algorithm performs two distinct stages at the arrival of each new task: (1) it pre-trains a policy on the prior experience stored in its replay buffer, and (2) it improves the policy with \emph{selective} data replay. Notably, both stages leverage previously collected data, and we find in an ablation study that each component serves an important role in our framework. Our simulated experiments consider two lifelong RL problems: a sequence of key-insertion tasks and a sequence of valve-turning tasks. On both problems, our approach considerably improves upon existing methods for sequential transfer, including Progressive Nets~\citep{rusu2016progressive} and DARC~\citep{eysenbach2020off}. Finally, we evaluate the efficacy of our approach on a sequence of ten challenging tasks with varying objectives and physical setups with a Franka Emika robot arm (see Fig.~\ref{fig:robot_tasks}). It requires only $100$K time-steps to learn ten tasks and achieves a $2$x improvement over learning each task from scratch. 

%% file: related_work.tex
\section{Related Work}
\label{sec:related_work}

Reinforcement learning has allowed robotic agents to autonomously learn an impressive array of skills~\citep{kober2013reinforcement,levine2016end}, from locomotion~\citep{kohl2004policy,tedrake2004applied,haarnoja2018learning,lee2020learning} to the manipulation of diverse objects~\citep{gu2017deep,kalashnikov2018qt,lee2019making,andrychowicz2020learning}. However, robotic RL setups are often only designed with the goal of solving an individual task in mind. As a result, the lifetime of these learning agents begins and ends with a single task, and each new task is to be learned from scratch. With the risks associated with physical interactions in the real world, this is not a practical approach for a robot to learn a diverse set of skills. Therefore, in the design of our algorithm, we emphasize data efficiency when solving a series of tasks.

In principle, if the agent leverages knowledge accrued from previously-solved tasks, then it should learn new ones more efficiently than from scratch~\citep{taylor2009transfer,lazaric2012transfer,wang2016learning,finn2017model,nagabandi2018learning,zhao2020meld,arndt2020meta,song2020rapidly}. In RL agents, this knowledge can be transferred through representations~\citep{rusu2016progressive,devin2017learning}, learned models~\citep{finn2017deep}, network weights~\citep{fernandez2006probabilistic,rusu2015policy}, or experiences~\citep{kaelblinglearning,taylor2008transferring,lazaric2008transfer,andrychowicz2017hindsight,tirinzoni2018importance,tao2020repaint}. Our work focuses on the transfer of weights and experiences, which we view as general and complementary forms of knowledge. Multi-task learning also aims to transfer knowledge across tasks but achieves this by learning the set of tasks together~\citep{parisotto2016actor,teh2017distral,hausman2018learning,yang2020multi,yu2020gradient}. This framework has allowed robots to learn a range of goal-based tasks~\citep{kaelblinglearning,andrychowicz2017hindsight} and other tasks that only differ in their reward functions, but may be less practical when each task has a unique physical setup. Unlike these prior works, we study the \emph{sequential} transfer learning problem where data can only be collected from the current task rather than in a round-robin fashion, and study how to efficiently solve a sequence of tasks on a real robot manipulator.

A common form of sequential transfer is to reuse the weights of a policy network by fine-tuning them to the new task~\citep{fernandez2006probabilistic,julian2020efficient} or distilling the learned behavior~\citep{levine2016end,rusu2015policy,parisotto2016actor,schmitt2018kickstarting}. While learned policies and value functions readily offer prior task information in a compact form, they become less useful as the optimal policies between tasks are less similar. On the other hand, the raw experience accumulated from earlier tasks cannot be immediately used to generate behavior---we have to expend computational resources to optimize a policy with this data for example. Additionally, not all experiences may be relevant to the target task, which can be addressed by prioritizing samples by their relevance~\citep{taylor2008transferring,lazaric2008transfer,tirinzoni2018importance,tao2020repaint,eysenbach2020off}. Despite these challenges, individual experience samples also represent the most complete as well as unprocessed form of knowledge from a task, and hence can be used in flexible ways. For example, they can be used to optimize auxiliary objectives~\citep{isele2018selective}, combat catastrophic forgetting~\citep{rolnick2018experience}, and accelerate learning of new tasks~\citep{andrychowicz2017hindsight,yin2017knowledge,tao2020repaint}. Nonetheless, none of these prior methods study lifelong learning of a sequence of physical robotic manipulation tasks. Our approach combines the strengths of weight and experience transfer to improve the data efficiency of sequential robotic task learning, and significantly outperforms prior methods in our experiments.

In many continual learning setups, the algorithm has a fixed storage budget, and must select existing samples to discard from its replay buffer. Prior work has proposed selection strategies such as reservoir sampling~\citep{rolnick2018experience,chaudhry2019tiny,buzzega2020dark,balaji2020effectiveness}, which uniformly samples a datapoint to discard, and maximizing measures like surprise~\citep{isele2018selective,sun2021information}, diversity~\citep{aljundi2019gradient,bang2021rainbow}, or coverage of the state space~\citep{isele2018selective}. Our work sidesteps this problem by assuming the ability to retain all previous samples, since many robot learning settings are bottlenecked more by the speed of physical data collection than by hard disk size. With modern compute and hard drives, it is often practical to accumulate a significant amount of data and indeed modern machine learning systems have been most successful when trained on broad datasets that are much larger than those typically used for training robots~\citep{krizhevsky2012imagenet}.

%% file: preliminaries.tex
\section{Preliminaries}
\label{sec:preliminaries}

\newcommand{\ssp}{\mathcal{S}}
\newcommand{\asp}{\mathcal{A}}
\newcommand{\st}{\mathbf{s}_t}
\newcommand{\stp}{\mathbf{s}_{t+1}}
\newcommand{\at}{\mathbf{a}_t}
\newcommand{\reb}{\mathcal{D}}

\newcommand{\task}{\mathcal{T}}
\newcommand{\bs}{\mathbf{s}}
\newcommand{\ba}{\mathbf{a}}

A task is defined by a Markov decision process with a continuous state space $\ssp$ and action space $\asp$. The next state $\bs' \in \ssp$ is determined by (unknown) dynamics $p(\bs' | \bs, \ba)$, and at each time step, the environment returns a scalar reward $r(\bs, \ba)$. The goal of standard RL is to acquire a policy $\pi(\ba | \bs)$ that maximizes the expected sum of rewards $\mathcal{J}(\pi) \coloneqq \sum_{t=1}^T \mathbb{E}_{(\st, \at) \sim \rho_\pi} \left[ \gamma^t r(\st, \at) \right],$ where $\rho_\pi$ is the trajectory distribution induced by $\pi$. Next, we introduce an RL algorithm that solves the single-task setting with off-policy experience (Sec.~\ref{subsec:sac}). To reuse experience from prior tasks, we measure the relevance of individual samples for a new task with importance weights (Sec.~\ref{subsec:importance_weighting}).

\subsection{Soft Actor-Critic}
\label{subsec:sac}
The soft actor-critic~\citep{haarnoja2018soft} (SAC) algorithm optimizes the maximum-entropy RL objective~\citep{ziebart2008maximum,toussaint2009robot}, using off-policy data to learn in a sample-efficient manner. The algorithm stores a replay buffer $\reb$ of all collected transitions and rewards, i.e., $(\bs, \ba, \bs', r)$. With this data, a Q-function (or critic) $Q_\theta$ is trained to minimize the Bellman error
$\mathcal{L}_Q = \mathbb{E}_{(\bs, \ba, \bs', r) \sim \reb} \left[ (Q_\theta (\bs, \ba) - (r + V(\bs')))^2 \right]$, where $V(\bs) = \mathbb{E}_{\ba \sim \pi(\cdot | \bs)} [Q_\theta(\bs, \ba) - \alpha \log \pi_\phi(\ba | \bs)]$
and $\alpha$ is a temperature parameter. The policy (or actor) $\pi_\phi$ is trained to minimize the KL divergence
$$\mathcal{L}_\pi = \mathbb{E}_{(\bs, \ba) \sim \reb} \left[  D_\text{KL} \left( \pi_\phi (\ba | \bs) \bigg|\bigg| \frac{\exp(Q_\theta(\bs, \ba))}{Z_\theta(\bs)} \right) \right]$$
where $Z_\theta$ is the partition function that normalizes the distribution $\exp(Q_\theta(\bs, \ba))$. In the overall algorithm, the agent alternates between collecting and storing data, and updating the actor and critic with this stored data. We build upon SAC as it is an effective RL algorithm that allows us to leverage previously collected data.

\subsection{Importance Weighting}
\label{subsec:importance_weighting}

Domain adaptation methods typically leverage importance weighting to correct the bias of samples from the source domain~\citep{zadrozny2004learning,baktashmotlagh2014domain}. When the tasks have different dynamics but same reward, prior work has defined the importance weights for each transition sample as the likelihood ratio $w(\bs, \ba, \bs') \coloneqq p^\text{tgt} (\bs' | \bs, \ba) / p^\text{src} (\bs' | \bs, \ba),$ where $p^\text{src}$ are the transition probabilities in the source task and $p^\text{tgt}$ are those in the target task. These weights can be estimated with learned probabilistic models~\citep{tirinzoni2018importance} or with classifiers in a likelihood-free manner~\citep{bickel2007discriminative,eysenbach2020off}. Eysenbach et al.~\citep{eysenbach2020off} use the estimated weights to relabel the rewards, i.e. $\tilde{r}(\bs,\ba,\bs') = r(\bs,\ba,\bs') + \log \hat{w}(\bs,\ba,\bs')$ in their method DARC, so that transitions that are likely under the target domain are weighed higher and vice versa. 

While our method will build upon DARC, DARC on its own is ill-suited for the lifelong learning setting for two reasons. First, this definition of the importance weight assumes that the state-action distributions are the same in the source and target domain datasets, i.e., $p^\text{src}(\bs, \ba) = p^\text{tgt}(\bs, \ba)$. DARC uses the same policy in the two domains, hence this approximately holds. However, we aim to learn a separate policy for each new task, making $p^\text{src}(\bs, \ba) / p^\text{tgt}(\bs, \ba)$ non-neglible. Second, the DARC setting places stronger limitations on the agent's access to the target domain, training the policy on \emph{all} of the source-domain data is imperative. However, our setting allows the agent to improve in the target domain collecting new data as necessary. Due to the inevitable estimation error in $\hat{w}$, we find that training on all the source domain data, even if re-weighted, can be counterproductive.

%% file: method.tex
\section{Lifelong Reinforcement Learning via Experience Transfer}
\label{sec:method}

The goal of our framework is to learn a series of robotic tasks in a practical and data-efficient manner. To this end, we describe the sequential multi-task learning problem (Sec.~\ref{subsec:problem}), in which the agent must learn policies for each task in sequence. We then devise an algorithm that leverages the prior experiences collected by the robot to improve sample efficiency when learning each additional task. 

\subsection{Sequential Multi-Task Learning Problem}
\label{subsec:problem}

We are interested in solving a sequence of tasks $\task^1, \task^2, \dots, \task^N$. Formally, each task $\task^i$ is defined by a MDP, with state space $\ssp$, action space $\asp$, transition dynamics $p^i(\bs' | \bs, \ba)$, and reward function $r^i(\bs, \ba)$. Our work focuses on transfer between tasks on the same robotic platform, and thus assumes the state-action space is shared between tasks. However, because each task potentially presents a different physical environment setup and objective, the dynamics and reward functions may differ. Each task has a success criterion that determines if the current policy has successfully solved the given task. This criterion is designed by the user and verified either manually by the user or automatically. Crucially, the robot is only given the next task when it has satisfied the success criterion of the current one or has spent the maximum number of attempts. 

\noindent \textbf{Comparison to continual reinforcement learning.} Unlike some continual RL formulations and approaches~\citep{rusu2016progressive,schwarz2018progress}, we assume in this setting that offline data from previous tasks can be stored and retrieved for future use~\citep{isele2018selective,rolnick2018experience,yin2017knowledge,tao2020repaint}. This work also focuses on maximizing the forward transfer performance and \emph{not} tackling catastrophic forgetting, a different challenge of continual learning. Also, motivated by the physical cost of switching tasks, we allow the robot to only collect new data in the task that it is currently solving.

\noindent \textbf{Access to reward functions.} Lastly, the environment is typically unknown to the agent. However, in robotics applications, the reward is often specified by the user, either directly provided as a closed-form function of states and actions, or indirectly as demonstrations of the task or examples of successful states. In the latter cases, a reward function can be recovered by inverse RL~\citep{ziebart2008maximum,wulfmeier2015maximum,finn2016guided,fu2017learning} or learning a classifier that distinguishes between successful and unsuccessful states~\citep{kalashnikov2018qt,fu2018variational,xie2018few,singh2019end}. Hence, we assume that the agent can access the full reward function for each task $r^i: \ssp \times \asp \to \mathbb{R}$, and not just the immediate reward feedback. As we describe next, this access allows us to \emph{relabel} old experience with the new reward function so they are consistent with the new task.

\begin{figure*}
\begin{minipage}[t]{0.5\linewidth}
\begin{algorithm}[H]
\small
\caption{Lifelong RL by Retaining Experiences}
\label{alg:full_algorithm}
\begin{algorithmic}[1]

\State $\reb^1, \theta^1, \phi^1 \gets \textsc{SAC}(\task^1)$ 
\For{$i=2, \dots, N$}
\State $\reb^\text{src}, \theta^i, \phi^i \gets \textsc{Pretrain}(\reb^{1:i-1}, r^i) $
\State $\reb^i, \theta^i, \phi^i \gets \textsc{Improve}(\reb^\text{src}, \theta^i, \phi^i)$

\EndFor
\end{algorithmic}
\end{algorithm}

\vspace{-0.5cm}

\begin{algorithm}[H]
\small
\caption{\textsc{Pretrain}}
\label{alg:pretrain}
\begin{algorithmic}[1]
\State \textbf{Input:} Replay buffers $\reb^1, \reb^2, \dots, \reb^{i-1}$
\State \textbf{Input:} Reward function $r^i$
\State \textbf{Optional:} Task parameters $\theta^{i-1}$, $\phi^{i-1}$

\State Initialize parameters $\theta^i$, $\phi^i$
\State Aggregate and relabel buffers (Eqn.~\ref{eq:1})

\For{each iteration}

\State Sample batch from $\reb^\text{src}$
\State Soft actor-critic updates:
\State $\diamond ~ \theta^i \gets \theta^i - \alpha \nabla_{\theta^i} \mathcal{L}_Q$
\State $\diamond ~ \phi^i \gets \phi^i - \alpha \nabla_{\phi^i} \mathcal{L}_\pi$

\EndFor
\State \textbf{Return:} $\reb^\text{src}, \theta^i, \phi^i$
\end{algorithmic}
\end{algorithm}

\end{minipage}
\begin{minipage}[t]{0.5\linewidth}
\begin{algorithm}[H]
\small
\caption{\textsc{Improve}}
\label{alg:finetune}
\begin{algorithmic}[1]
\State \textbf{Input:} Relabeled source buffer $\reb^\text{src}$
\State \textbf{Input:} Pre-trained parameters $\theta^i$, $\phi^i$
\State Initialize replay buffer $\reb^i$
\State Initialize classifier $\psi$

\For{each iteration}

\For{each environment step}
\State Sample action $\ba \sim \pi_\theta(\ba | \bs)$
\State Step in environment $\bs' \sim p^i(\bs' | \bs, \ba)$
\State Update buffer $\reb^i \leftarrow \reb^i \cup \{ (\bs, \ba, \bs', r) \}$
\EndFor

\For{each update step}

\State Sample batch from $\reb^\text{src} \cup \reb^i$
\State Soft actor-critic updates:
\State $\diamond ~ \theta^i \gets \theta^i - \alpha \nabla_{\theta^i} \mathcal{L}_Q$
\State $\diamond ~ \phi^i \gets \phi^i - \alpha \nabla_{\phi^i} \mathcal{L}_\pi$

\State Classifier update:
\State $\diamond ~ \psi \gets \psi - \alpha \nabla_{\psi} \mathcal{L}_{\psi}$

\EndFor

\State Filter source buffer $\reb^\text{src}$ (Eqn.~\ref{eq:2})

\EndFor
\State \textbf{Return:} $\reb^i$, $\theta^i$, $\phi^i$
\end{algorithmic}
\end{algorithm}
\end{minipage}
\vspace{-.2cm}
\end{figure*}

\subsection{Algorithm Overview}
\label{subsec:overview}

Our algorithm, depicted in Fig.~\ref{fig:framework} and summarized in Alg.~\ref{alg:full_algorithm}, begins by learning the first task from scratch with standard SAC. For each following new task, we restore the replay buffer(s) from the previous task(s) and pre-train on this data (Sec.~\ref{subsec:pretrain}). We then improve the agent's policy on the task with online experience (Sec.~\ref{subsec:improve}). Note that both of these stages reuses the previously collected, offline data in order to maximize the sample efficiency of our algorithm.

\subsection{Pre-Training on Prior Experience}
\label{subsec:pretrain} 

At the arrival of a new task $\task^i$, we can either (1) randomly initialize the actor and critic weights or (2) restore them from the previous task. In our evaluation of both variants (Sec.~\ref{sec:experiments}), we find that depending on the similarity of the tasks, transferring the weights can result in better or worse performance compared to random initialization. Irrespective of the choice of initialization, we next restore the replay buffers $\reb^{1:i-1}$ from the previous tasks. Then, with the reward function $r^i$ for task $\task^i$, we can relabel the rewards of the aggregated dataset, making them consistent with task $\task^i$
\begin{equation} \label{eq:1}
\begin{split}
\reb^\text{src} &\coloneqq \bigcup_{j=1}^{i-1} \{ (\bs, \ba, \bs', r^i(\bs, \ba)) ~|~ (\bs, \ba, \bs', r) \in \reb^j \}.
\end{split}
\end{equation}
We subsequently sample batches of this modified dataset $\reb^\text{src}$ with which we apply offline updates to the policy and critic as a form of pre-training. Despite the potential discrepancies in the dynamics between tasks, we expect the pre-trained parameters to produce better trajectories than a random policy or even the policy from the previous task, and thus accelerate learning in the new task. The pre-training subroutine is summarized in Alg.~\ref{alg:pretrain}. 

\subsection{Improving with Online Experience}
\label{subsec:improve}

To improve the pre-trained policy from Sec.~\ref{subsec:pretrain}, our algorithm next collects online interactions in the robot's physical learning environment for task $\task^i$. Since our goal is to minimize the amount of online experience the robot needs to collect in its environment, we aim to also use the relabeled experience $\reb^\text{src}$ during this online phase. However, because of the differing dynamics between tasks, these samples may vary in utility for accomplishing the current task. Addressing the dynamics gap is significantly more challenging, because unlike the reward function which is commonly specified by the user, we rarely have access to a model of the environment. One approach is to learn to model the dynamics $p^i(\bs' | \bs, \ba)$ and relabel the transitions by querying the model. However, increasingly complex dynamics demand more expressive models and thus more environment samples to fit them. It is often easier and more efficient to instead estimate the importance weights, i.e., the likelihood ratio of samples under the target task versus the source tasks.

\noindent \textbf{Likelihood-free importance weights.} Let $\reb^\text{tgt}$ represent the samples from the target task $i$ and $\reb^\text{src}$ samples from all previous tasks. Our algorithm trains a separate policy for each new task, leading to unequal state-action distributions $p^\text{tgt}(\bs, \ba) \not = p^\text{src}(\bs, \ba)$. Hence, we define the importance weights as $w(\bs, \ba, \bs') = p^\text{tgt}(\bs, \ba, \bs') / p^\text{src} (\bs, \ba, \bs')$. In contrast to the importance weights in DARC~\citep{eysenbach2020off} and in off-policy RL, the $w$ we define here account and correct for shifts in both the transition dynamics due to the different tasks, and the marginal state-action distribution due to the different policies. To estimate $w$ for samples $(\bs, \ba, \bs') \in \reb^\text{src}$, we first express the likelihood ratio with Bayes' rule as
\begin{align*}
    w(\bs,\ba,\bs') &= \frac{p(\text{target} | \bs, \ba, \bs')}{p(\text{source} | \bs, \ba, \bs')} \cdot \frac{p(\text{source})}{p(\text{target})}.
\end{align*}
This expression is a more useful form since we can estimate the first term with a classifier $c_\psi (\bs, \ba, \bs')$ that outputs the probability that a $(\bs, \ba, \bs')$ tuple is from the target task, trained with the cross-entropy loss:
\begin{align*}
\mathcal{L}_{\psi} &= -\mathbb{E}_{(\bs, \ba, \bs') \sim \reb^i} \left[ \log c_{\psi} (\bs, \ba, \bs') \right] - \mathbb{E}_{(\bs, \ba, \bs') \sim \reb^\text{src}} \left[ \log (1- c_{\psi} (\bs, \ba, \bs')) \right].
\end{align*}
The second term can be estimated with the ratio of the replay buffers' size, $|\reb^\text{src}| / |\reb^i|$. Intuitively, the $\reb^\text{src}$ samples weigh less as we collect more samples in the target task.

\noindent \textbf{Which samples should we transfer?} One way to use these weights is to re-weigh the examples in the RL objective. However, the weights can be numerically unstable and may require clipping to lie in a more reasonable range. 
We instead filter out samples that are unlikely in the target task, according to the classifier $c_\psi$. Concretely, we apply a threshold to the first term of the importance weight:
\begin{align*}
    \tilde{w}(\bs, \ba, \bs') &= \mathds{1}\left( \frac{c_\psi (\bs, \ba, \bs')}{1 - c_\psi (\bs, \ba, \bs')} \ge \gamma \right)
\end{align*}
Thresholding allows us to control how conservatively samples are transferred with $\gamma$. We incorporate the second term by sampling training batches according to the ratio $|\reb^\text{src}| / |\reb^i|$. As $|\reb^\text{src}|$ remains fixed and $|\reb^i|$ grows, we in turn upsample from the target task buffer $\reb^i$ and downsample from the source buffer $\reb^\text{src}$. Our overall filtering rule is:
\begin{equation} \label{eq:2}
\reb^\text{src} \gets \left\{ (\bs, \ba, \bs', r)  ~\Bigg|~ (\bs, \ba, \bs', r) \in \reb^\text{src},  \frac{c_{\psi}(\bs, \ba, \bs')}{1-c_{\psi}(\bs, \ba, \bs')} \ge \gamma \right\}
\end{equation}
We train the classifier $c_{\psi}$ online and re-filter the source data $\reb^\text{src}$ at regular intervals. The complete online improvement phase is outlined in Alg.~\ref{alg:finetune}.

%% file: experiments.tex
\section{Experiments}
\label{sec:experiments}

\newcommand{\both}{\textbf{Ours (warm-start)}}
\newcommand{\data}{\textbf{Ours}}

Our experiments seek to answer the following questions: \textbf{(1)} How does our approach compare to existing methods for sequential transfer? 
\textbf{(2)} How important is the pre-training phase and how important is filtering prior data in the online fine-tuning phase?
\textbf{(3)} Can our approach sequentially learn to solve a series of manipulation tasks on a robot? 
The videos accompanying our experiments are on our project webpage: \url{https://sites.google.com/view/retain-experience/}.

\subsection{Experimental Setup}
\textbf{Comparisons.} To answer question \textbf{(1)}, we carefully study our method as well as several comparisons in a simulated robot environment. In particular, we compare to two methods that leverage previously trained weights:
\begin{itemize}[leftmargin=*]
    \item \textbf{Progressive Nets}~\citep{rusu2016progressive}. Continual learning framework that transfers previously learned features.
    
    \item \textbf{Fine-tuning}~\citep{julian2020efficient}. Ablation of our method that restores the weights from the previous task.
\end{itemize}
We also compare to methods that reuse previously collected data:
\begin{itemize}[leftmargin=*]
    \item \textbf{DARC}~\citep{eysenbach2020off}. Domain adaptation method that defines the importance weights $w^\text{DARC}(\bs, \ba, \bs') = p^\text{tgt}(\bs' | \bs, \ba) / p^\text{src}(\bs' | \bs, \ba)$, which are estimated with two classifiers $c(\bs, \ba, \bs')$ and $c(\bs, \ba)$, and adds $\Delta r = \log w^\text{DARC}$ to the rewards of prior data. DARC is designed to only learn from data collected in the source domain. To use both old and new data, we only relabel the experiences from previously solved tasks, and clip $\Delta r$ from above at $0$. We find these modifications improve the performance of DARC.

    \item \textbf{Off-policy IW}. Vanilla off-policy RL that reweighs samples with the importance weights $w^\text{OP} = \pi^i(\ba | \bs)/ \pi^{i-1}(\ba | \bs)$.
\end{itemize}
Finally, we evaluate our approach and its ablations:
\begin{itemize}[leftmargin=*]
    \item \textbf{Ours}. The weights of this variant are randomly initialized before the pre-training phase of each task, and only the replay buffers are transferred.
    
    \item \textbf{Ours (warm-start)}. In addition to the replay buffers, we also restore the trained weights of the policy and critic from the previous task.
    
    \item \textbf{Ours (with DARC weights)}. An ablation of \textbf{Ours} that does not use the filtering scheme introduced in Section~\ref{subsec:improve} and instead, like DARC, relabels the rewards as $r + \log w^\text{DARC}$. 
\end{itemize}

\captionsetup[subfloat]{captionskip=5pt}

\begin{figure}
    \centering
    \begin{minipage}{\linewidth}
    \centering
    \subfloat[]{
        \includegraphics[height=1.1in]{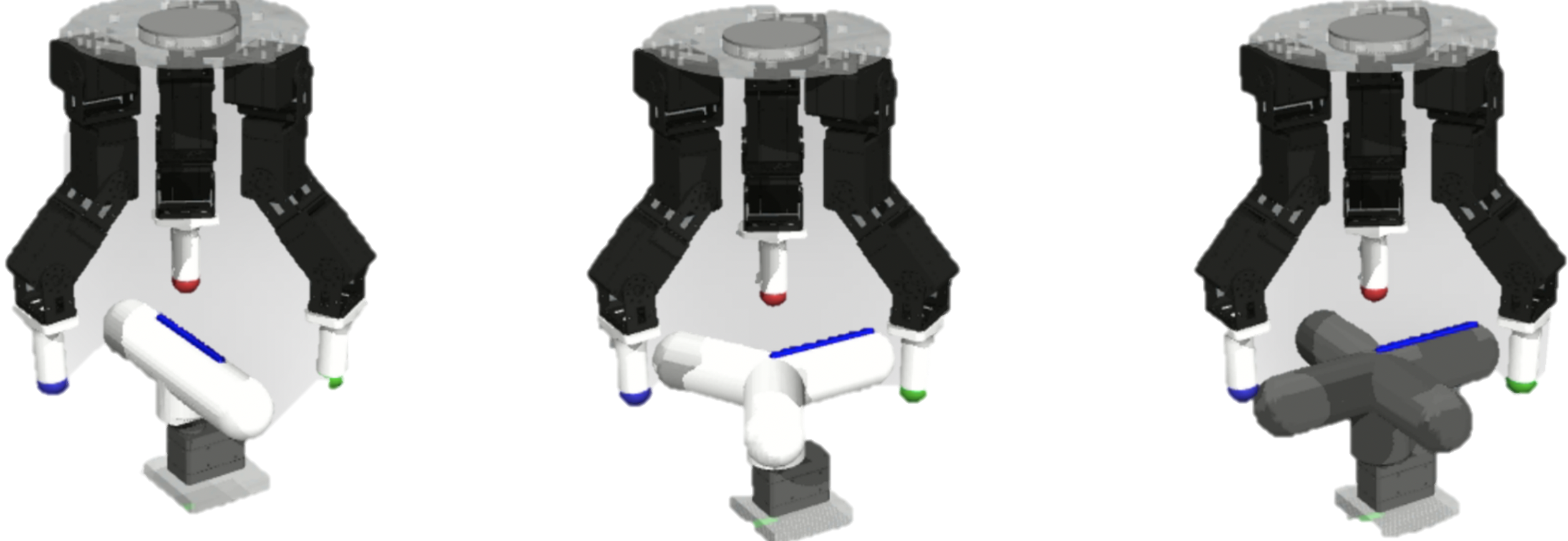}
        \label{fig:dclaw_tasks}
    }
    \hspace{0.7cm}
    \subfloat[]{
        \includegraphics[height=1in]{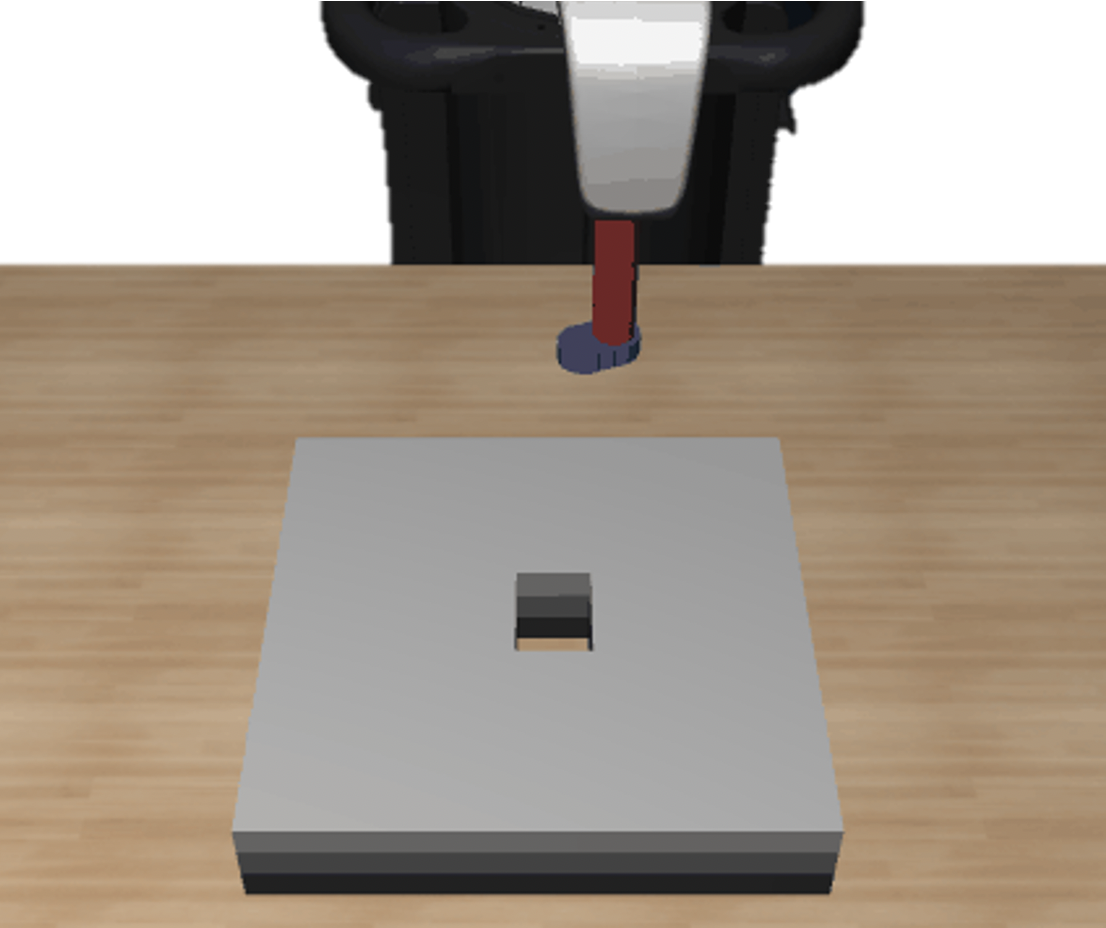}
        \hspace{0.5cm}
        \includegraphics[height=1in]{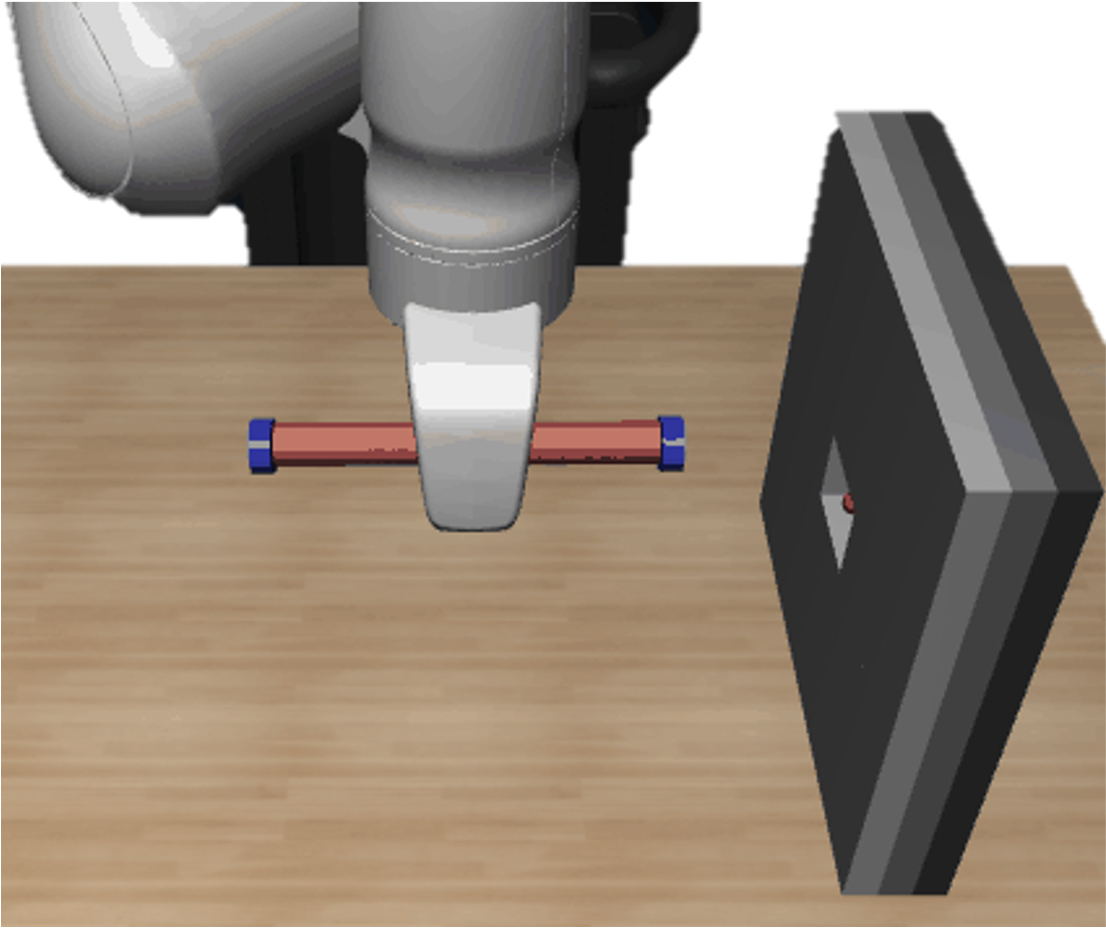}
        \label{fig:key_tasks}
    }
    \caption{\small (a) D'Claw tasks with different valve shapes (image taken from~\citep{yang2022evaluations}). (b) Examples of vertical and horizontal key-insertion tasks.
    }
    \label{fig:simulated_tasks}
    \end{minipage}
\end{figure}

\textbf{Simulated environments.} We evaluate each method on two simulated robotic environments, depicted in Fig.~\ref{fig:simulated_tasks}. In each, the algorithm must learn to solve a sequence of tasks, with varying dynamics and reward functions.
\begin{itemize}[leftmargin=*]
    \item \textbf{ROBEL D'Claw}~\citep{yang2022evaluations}. This benchmark includes a set of $10$ valve-rotation tasks, each with a different valve shape. The tasks also differ in the position feedback gain, friction coefficients, Cartesian position of the valve, target orientation of the valve, and desired direction of rotation, i.e., clockwise versus counterclockwise. We design \emph{random} task sequences of $10$ tasks each, which have randomly sampled parameters defining each task. For visualization, we additionally design a \emph{hard} task sequence that alternates between rotating the valve clockwise and counterclockwise, in addition to randomizing the other task parameters.
    
    \item \textbf{Key Insertion}. We construct a family of key-insertion tasks within Robosuite~\citep{robosuite2020}, using the MuJoCo physics engine~\citep{mujoco}. The objective of each task is to insert the key into a box, while different tasks have varying placements of the box, key sizes, and initial orientations of the key with respect to the box. For more task heterogeneity, we rotate the box in a subset of the tasks, and the key has to be horizontally inserted, rather than vertically. We again design \emph{random} task sequences of $16$ tasks, each with randomly sampled task parameters.
    Similar to the D'Claw domain, we design a \emph{hard} task sequence that alternates between vertical and horizontal insertion tasks, in addition to randomizing the other parameters.
\end{itemize}
A full description of each environment is provided in App.~\ref{app:envs}.

\textbf{Evaluation metrics.} In our evaluation of each method, we track the \textbf{average performance} across the agent's entire lifetime and the \textbf{final performance} on each task of the sequence after $T_\text{task}$ time-steps. In our simulations, the performance is measured in terms of task success, and the agent trains for $T_\text{task} = 50$k steps in each task.

\subsection{Evaluating Forward Transfer}
\label{subsec:forward_results}

\noindent \textbf{Comparative evaluation.} In Table~\ref{tbl:sim_results}, we summarize the performance of all methods in terms of task success. In the ROBEL D'Claw domain, our method outperforms all comparisons in terms of average and final task success, matching the performance of learning each task from scratch with $100$K time-steps. On average, the final performance of our method on each task is $82\%$. Critically, \textbf{Ours (warm-start)}, which additionally transfers the policy and critic network weights, is less successful, indicating that this initialization harms performance. We hypothesize that the transfer of weights is less useful due to the heterogeneity in the tasks, specifically since tasks can vary in which direction the valve needs to be rotated. Nonetheless, we see that \textbf{Ours (warm-start)} improves upon naive fine-tuning, hence demonstrating the benefits of our data transfer scheme. \textbf{Progressive Nets} notably leverages the previously trained weights the most constructively, as it achieves the highest average and final task success of all weight transfer methods. 

The data transfer methods, \textbf{DARC} and \textbf{Off-Policy IW}, perform worse than if the agent learned each task from scratch with the same number of online samples. However, augmenting DARC with our pre-training scheme significantly improves its performance. In the Key Insertion tasks, both of the data transfer methods perform competitively with our approach, which suggests that the exact differences in the importance weights are less important here. However, \textbf{Ours} is the only method that achieves both high average and high final task performance. 

\begin{table}{}
    \vspace{-0.2cm}
    \centering
    \small
    \setlength{\tabcolsep}{11pt}
    \begin{tabularx}{0.85\linewidth}{lcccc}
        \toprule
         & \multicolumn{2}{c}{ROBEL D'Claw} &  \multicolumn{2}{c}{Key Insertion} \\
        Method & Average Perf & Final Perf & Average Perf & Final Perf \\
        \cmidrule(lr){1-1} \cmidrule(lr){2-3} \cmidrule(lr){4-5} 
        Scratch (50k) & $0.28 \pm 0.02$ & $0.59 \pm 0.03$ & $0.26 \pm 0.02$ & $0.52 \pm 0.04$ \\
        \textcolor{gray}{Scratch (100k)} & \textcolor{gray}{$0.47 \pm 0.02$} & \textcolor{gray}{$0.85 \pm 0.03$} & \textcolor{gray}{$0.45 \pm 0.03$} & \textcolor{gray}{$0.69 \pm 0.04$} \\
        \cmidrule(lr){1-1} \cmidrule(lr){2-3} \cmidrule(lr){4-5} 
        Prog. Nets & $0.41 \pm 0.02$ & $0.74 \pm 0.02$ & $0.50 \pm 0.04$ & $0.75 \pm 0.03$ \\
        Fine-tuning & $0.27 \pm 0.02$ & $0.50 \pm 0.04$ & $0.49 \pm 0.03$ & $0.63 \pm 0.03$ \\
        \cmidrule(lr){1-1} \cmidrule(lr){2-3} \cmidrule(lr){4-5} 
        DARC & $0.23 \pm 0.01$ & $0.56 \pm 0.03$ & $0.60 \pm 0.02$ & $\mathbf{0.89 \pm 0.02}$ \\
        Off-policy IW & $0.33 \pm 0.04$ & $0.59 \pm 0.04$ & $\mathbf{0.69 \pm 0.03}$ & $0.82 \pm 0.03$ \\
        \cmidrule(lr){1-1} \cmidrule(lr){2-3} \cmidrule(lr){4-5} 
        \textbf{Ours} & $\mathbf{0.49 \pm 0.02}$ & $\mathbf{0.82 \pm 0.02}$ & $\mathbf{0.69 \pm 0.03}$ & $\mathbf{0.87 \pm 0.03}$ \\
        \textbf{Ours (warm-start)} & $0.32 \pm 0.05$ & $0.61 \pm 0.07$ & $0.50 \pm 0.03$ & $0.62 \pm 0.04$ \\
        \textbf{Ours (w. DARC weights)} & $0.41 \pm 0.03$ & $0.77 \pm 0.03$ & $\mathbf{0.66 \pm 0.01}$ & $0.76 \pm 0.03$ \\
        \bottomrule
      \end{tabularx}
    \vspace{-0.1cm}
    \caption{\small The average and final task success of each method on the ROBEL D'Claw and Key Insertion domains. In each simulated domain, we evaluate on $5$ \emph{random} task sequences. We report the mean and standard error in each entry.}
    \label{tbl:sim_results}
    \vspace{-0.3cm}
\end{table}

\begin{figure}
    \centering
    \includegraphics[width=\linewidth]{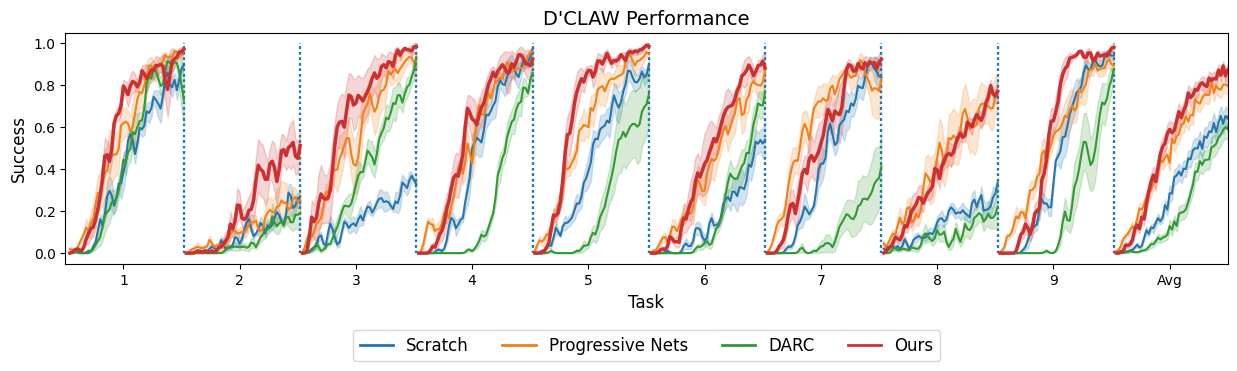}
    \includegraphics[width=\linewidth]{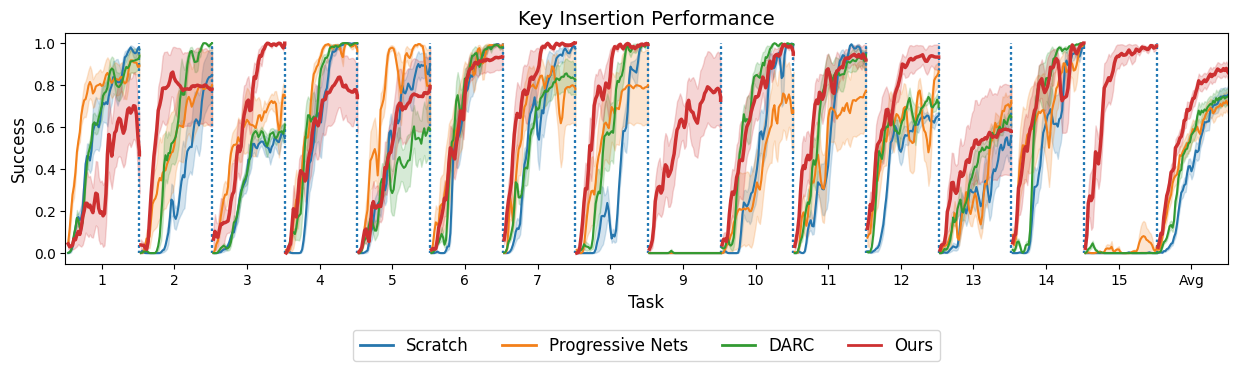}
    \caption{\small Lifetime performance of our method and select baselines on the \emph{hard} task sequence of each domain. Shaded regions depict the standard error across $5$ random seeds, and the solid lines depict their average. The final curve averages the learning curves across all of the tasks in the sequence.}
    \label{fig:hard_seq}
\end{figure}

In Fig.~\ref{fig:hard_seq}, we visualize the lifetime performance of each method on the \emph{hard} task sequence of both simulated domains. Of the baselines, we select \textbf{Scratch}, \textbf{Progressive Nets}, and \textbf{DARC} to visualize. The latter two are chosen as the representative weight transfer and data transfer approaches. We evaluate these methods on the two sequences and plot the average over $5$ trials each with a different random seed. On the D'Claw sequence, our approach has a final performance of $87\%$, with a slight advantage over \textbf{Progressive Nets}, which has a success rate of $80\%$. In the Key Insertion domain, however, our approach on average achieves a final success rate of $87\%$ while \textbf{Progressive Nets} and \textbf{DARC} perform comparably to learning from scratch.

\noindent \textbf{Improved efficiency.} A desirable property in sequential learning is compounding learning, that is, improved data efficiency as more tasks are seen. We evaluate this property on a held-out horizontal insertion task after training on a varying number of tasks, and summarize the results in terms of the final task success in Table~\ref{tbl:improved_efficiency}. Generally, the task success trends upwards as we increase the number of prior tasks we train on, suggesting that the data efficiency of our algorithm improves as we provide more tasks. We hypothesize that the performance improves with more tasks because of the heterogeneity in our task sequence, i.e.,
\begin{wraptable}{r}{0.5\linewidth}
    \vspace{-0.3cm}
    \centering
    \small
    \setlength{\tabcolsep}{1em}
    \begin{tabularx}{\linewidth}{lcccc}
        \toprule
        \# of Total Tasks & $1$ & $3$ & $5$ & $7$ \\
        \# of Horizontal Tasks & $0$ & $0$ & $1$ & $2$ \\
        \midrule
        Ours & $0.48$ & $0.54$ & $0.81$ & $0.84$ \\
        \bottomrule 
    \end{tabularx}
    \vspace{-0.2cm}
    \caption{\small Average task success 
    after learning a varying number of tasks. 
    The performance trends upwards with more tasks.
    }
    \vspace{-0.4cm}
    \label{tbl:improved_efficiency}
\end{wraptable}
having both horizontal (H) and vertical (V) insertion tasks. In particular, when $N=1$ and $N=3$, all the prior tasks are V tasks. When $N=5$, the first H task is introduced, leading to better transfer on the target task, which is also an H task. The introduction of the second H task in the $N=7$ case leads to another improvement, albeit a smaller one.

\subsection{Ablations: Investigating Advantages of Prior Experience}
The results from Sec.~\ref{subsec:forward_results} demonstrate that the previously collected data samples, when appropriately leveraged, are a powerful form of knowledge transfer. To answer question \textbf{(2)} and to verify our algorithm utilizes this prior experience more effectively than alternative design choices, we ablate the pre-training and online improvement stages of our framework. The ablations are evaluated on $5$ random target tasks.

\begin{figure*}[t]
    \centering
    \begin{minipage}{\linewidth}
    \centering
    \subfloat[]{
        \includegraphics[height=1.6in]{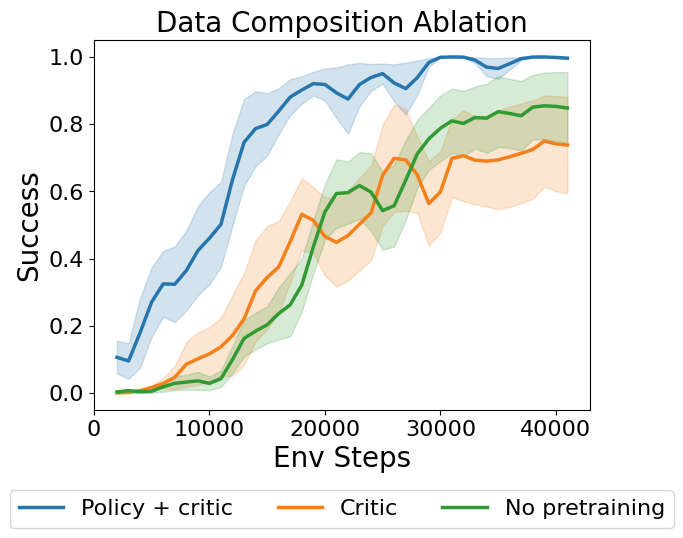}
        \label{fig:ablations_pretrain}
    } 
    \subfloat[]{
        \includegraphics[height=1.6in]{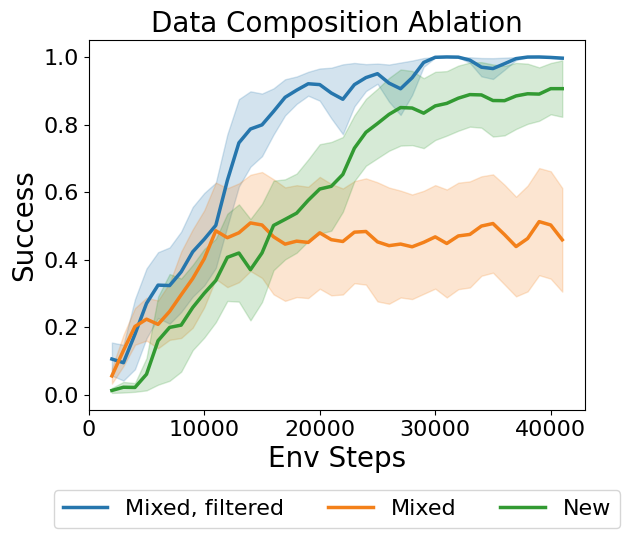}
        \label{fig:ablations_improve}
    }
    \subfloat[]{
        \includegraphics[height=1.6in]{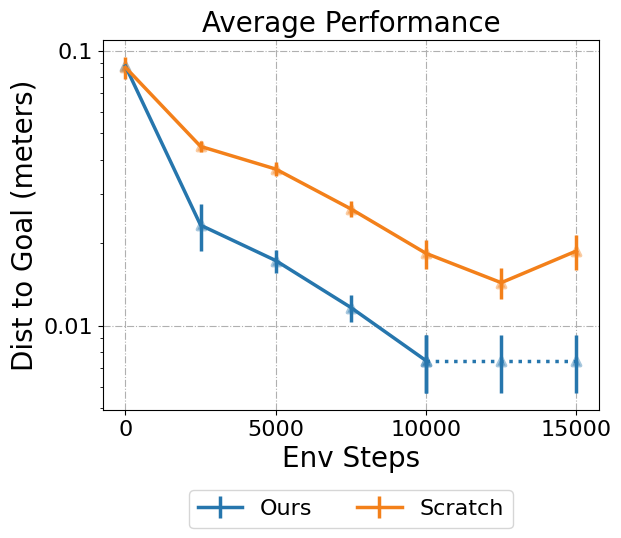}
        \label{fig:robot_results}
    }
    \vspace{-.2cm}
    \caption{\small (a-b) Ablations of the pre-training and online phases, evaluated on $5$ different target tasks. Shaded regions indicate the standard error across the $5$ trials; the solid lines represent their average. (c) The learning curve averaged across tasks $i>1$ on the physical robot. Each data-point represents the distance to goal at the final time-step across $10$ trials, averaged across the $9$ tasks. 
    The error bars represent $95\%$ CI. 
    }
    \label{fig:ablations}
    \end{minipage}
\end{figure*}

\noindent \textbf{Pre-training.} Our algorithm pre-trains both the policy and critic with the relabeled data from $\reb^\text{src}$ as its first step of learning a new task. Alternatively, we can pre-train the critic only or not pre-train at all. In this comparison, we randomly initialize all weights prior to pre-training to isolate its effects. The results, averaged across $5$ different target tasks for each method, are presented in Fig.~\ref{fig:ablations_pretrain}, and suggest that pre-training both the policy and critic weights leads to more efficient learning, compared to only pre-training the critic weights and no pre-training.

\noindent \textbf{Online improvement.} In the online improvement step, our algorithm trains on a mixture of filtered prior experience and new experience collected online. We compare this data composition to (1) uniformly mixing the prior and new experience, akin to standard experience replay, and (2) training on new experience only. For each variant, we first randomly initialize the policy and critic weights and perform the pre-training phase. Then, in the online fine-tuning phase, we evaluate each of the three data composition schemes. As shown in Fig.~\ref{fig:ablations_improve}, filtering out the prior samples that are unlikely under the current dynamics leads to improved data efficiency. In fact, the standard experience replay scheme finds a suboptimal policy as a result of training on unfiltered data from previous tasks. 

\subsection{Learning in the Real World}
\label{sec:realworld}

To answer question \textbf{(3)}, we evaluate our approach with a Franka Emika Panda robot arm on a sequence of challenging tasks with varying physical setups.

\noindent \textbf{Experimental setup.} We evaluate our algorithm on a physical robot arm on a sequence of $10$ object manipulation tasks, ranging from capping a bottle to inserting a block (see Fig.~\ref{fig:robot_tasks}). The robot's state consists of its end-effector pose, and takes actions corresponding to $3$D Cartesian end-effector displacements at $5$ Hz. We evaluate \data, as it outperformed the other variants in the simulations.

\begin{table*}
    \centering
    \setlength{\tabcolsep}{4pt}
    \scriptsize
    \begin{tabularx}{\linewidth}{l*{10}{c}}
        \toprule
        Task & $1$ & $2$ & $3$ & $4$ & $5$ & $6$ & $7$ & $8$ & $9$ & $10$ \\
        \midrule
        Scratch & $1.11$\tiny$\pm 0.05$ & $2.65$\tiny$\pm 0.24$ & $1.62$\tiny$\pm 0.02$ & $1.27$\tiny$\pm 0.09$ & $0.83$\tiny$\pm 0.15$ & $2.81$\tiny$\pm 0.23$ & $1.90$\tiny$\pm 0.03$ & $2.11$\tiny$\pm 0.28$ & $1.30$\tiny$\pm 0.23$ & $1.99$\tiny$\pm 0.19$  \\
        \data & -- & $\mathbf{0.96}$\tiny$\mathbf{\pm 0.05}$ & $\mathbf{1.30}$\tiny$\mathbf{\pm 0.25}$ & $\mathbf{0.66}$\tiny$\mathbf{\pm 0.08}$ & $\mathbf{0.53}$\tiny$\mathbf{\pm 0.05}$ & $\mathbf{0.57}$\tiny$\mathbf{\pm 0.01}$ & $\mathbf{0.50}$\tiny$\mathbf{\pm 0.01}$ & $\mathbf{0.81}$\tiny$\mathbf{\pm 0.22}$ & $\mathbf{0.78}$\tiny$\mathbf{\pm 0.06}$ & $\mathbf{0.60}$\tiny$\mathbf{\pm 0.01}$ \\
        \bottomrule
    \end{tabularx}
    \caption{\small The final distance to goal (in centimeters) for each of the $10$ tasks on the Franka robot. We evaluate the policies after $10$K environment steps for $10$ trials, and report the means and standard errors.}
    \label{tbl:robot_results_table}
\end{table*}

\noindent \textbf{Results.} We focus our evaluation on forward transfer and compare the learning efficiency of our algorithm to learning each task from scratch. After each epoch, we roll out the mean policy for $10$ evaluation episodes, and plot the average distance to the goal position versus the number of training environment steps averaged across all tasks after the first one in Fig.~\ref{fig:robot_results}. 
The individual learning curves for each task are included in App.~\ref{app:results}.
Overall, our algorithm achieves an average distance of $0.75$ centimeters to the goal within $10$K environment steps of a new task compared to an average distance of $1.83$ centimeters achieved by learning from scratch in the same number of steps. In Table~\ref{tbl:robot_results_table}, we report the average final performance by task for the two methods. Like our simulation results, compared to learning from scratch, our algorithm learns each task more efficiently by leveraging the retained experiences. Further, we qualitatively observe that one key benefit of retaining and relabeling the data is that it affords the agent fewer exploration steps in the regions of the state space that are irrelevant for the task at hand. 

%% file: conclusion.tex
\section{Discussion}
\label{ref:conclusion}

Algorithms for lifelong learning should allow robots to solve new tasks in succession by accumulating and building upon prior experience. To this end, we presented a simple framework that allows robots to learn sequences of tasks in a data-efficient manner by retaining and reusing prior experience. Our approach partitions the learning of each task into two stages: first, it pre-trains on relabeled prior data; second, it improves with online experience. In our experiments, we demonstrated this transfer of experience as a powerful mechanism for lifelong learning on physical robots.

Nonetheless, limitations of this approach remain, which we hope to address in future work. First, our algorithm assumes that all the previously collected data can be stored, which may present scalability challenges as the number of tasks to learn grows. However, we expect that naive approaches to discard experiences, e.g. uniformly by task, may be effective towards mitigating this challenge. Finally, the algorithm is in principle applicable to any task that can be solved with reinforcement learning; but, our experiments consider only reaching, placing, and insertion style tasks. We hope to study a wider range of manipulation tasks, which may require the use of visual observations or force feedback. We can also expect transfer to become more difficult as the tasks become more dissimilar, an interesting challenge for future investigation.

%% file: appendix.tex
\clearpage
\newpage
\appendix

\section{Hyperparameter Details}
\label{app:hyperparameters}

\noindent \textbf{Policy and critic networks.} For all experiments, we implement our algorithm on top of the soft actor-critic (SAC)~\cite{haarnoja2018soft} algorithm. The policy and critic are each MLPs with $2$ fully-connected layers of size $256$ and ReLU non-linearities.

\noindent \textbf{Domain classifier networks.} For all experiments, the domain classifier networks $D_1$, $D_2$ are each MLPs with $2$ fully-connected layers of size $256$ and ReLU non-linearities. Following~\cite{eysenbach2020off}, we inject Gaussian input noise with $\sigma = 1.0$ to combat overfitting at the beginning when there are few samples from the task currently being learned.

\noindent \textbf{Learning rates.} For our simulated experiments, we use the Adam optimizer and learning rate of $3\mathrm{e}{-4}$ for the policy and critic updates, and a learning rate of $1\mathrm{e}{-3}$ for the domain classifiers.
For our robot experiments, we use a learning rate of $1\mathrm{e}{-3}$ for the policy, critic, and domain classifier updates.

\noindent \textbf{Pre-training phase.} For all experiments, we pre-train the policy and critic of~\data~and~\both~with the relabeled data from the restored replay buffers for $10$k iterations before online improvement.

\noindent \textbf{Online improvement phase.} In the online improvement phase of~\data~and~\both, we use threshold values $\zeta_1 = 0.5$ and $\zeta_2 = 0.9$ for all experiments. We re-filter the source dataset $\mathcal{D}^\text{src}$ after every $1000$ iterations. At the beginning of the online phase, the batches used for the policy and critic updates are composed of $50\%$ filtered prior data and $50\%$ new online data. We increase this ratio $\rho$ of new data to prior data according to:
$$
\rho = \text{clip}(|\mathcal{D}^i| + 12500) / 25000, 1.0)
$$
where $|\rb^i|$ is the size of the replay buffer for the current task. In other words, the ratio is increased linearly and reaches $1.0$ once $25$k steps have been taken in the current task.

\section{Environment Setup}
\label{app:envs}

\subsection{Simulated Experiments}
In the experiments on the D'Claw benchmark~\citep{yang2022evaluations}, the objective is to rotate the valve to a target angle. Across tasks, we vary the valve shape, position feedback gain, friction coefficients, position of the value, target angle, desired direction of rotation. The reward function across all tasks is:
\begin{align*}
    -0.5 \cdot |s_\theta - g_\theta| + \mathbbm{1}(|s_\theta - g_\theta| < 0.05),
\end{align*}
where $s_\theta$ and $g_\theta$ are the valve's current and target angles respectively.

In the simulated key insertion experiments, we use the Robosuite~\cite{robosuite2020} simulation framework which employs the MuJoCo physics engine~\cite{mujoco}. The robot's state includes the robot's joint positions and velocities, its end-effector pose, and a binary indicator of whether the key is inside the robot's gripper. The action controls the deltas in the 3D-position and the z-axis rotation of the robot's end-effector. Between tasks, we vary the $xy$-position of the box $g_{xy}$, the relative orientation of the key to the hole $g_\theta$, and the length of the key $l$. The reward function across all tasks is:
\begin{align*}
    \mathbbm{1}(\| \bs_{xy} - g_{xy} \|_2 \le 0.03) \cdot \mathbbm{1}(\bs_z - l \le g_{z,u} + 0.005) - \tanh(\| 10 \cdot (\bs_{xy} - g_{xy} \|_2 + |\bs_z - g_{z,l}|)) - \tanh(|\bs_\theta - g_\theta|),
\end{align*}
where $\bs_{xy}$ is the $xy$-coordinates of the robot end-effector, $\bs_\theta$ is the $z$-axis rotation of the end-effector, $g_{z,u}$ is the $z$-coordinate of the top of the box, and $g_{z,l}$ is the z-coordinate of the bottom.
Note, in Fig.~\ref{fig:simulated_tasks}, that the box is composed of three ``layers,'' each colored with a different shade of gray. We measure task success on a scale of $\{0, 1, 2, 3 \}$ based on which layer the key head reaches at the final time-step of the episode, and report all results in terms of the normalized scores by dividing by $3$.

\subsection{Robot Experiments}
In our robot experiments, we design a series of $10$ tasks with varying setups and objectives. We describe these tasks below.

\noindent \textbf{Task 1: Reaching.} The robot is to reach a fixed goal position $g_\text{reach} = \begin{bmatrix} 0.466, 0.028, 0.153 \end{bmatrix}^T$ without any obstacles. The reward function for this task is
\begin{align*}
    r^1(\bs, \ba) &= 1 - \tanh(10 \cdot \| \bs_{xyz} - g_\text{reach} \|_2),
\end{align*}
where $\bs_{xyz}$ are the 3D Cartesian coordinates of the robot end-effector.

\noindent \textbf{Task 2: Marker insertion (A).} The objective is to align the marker to the corresponding hole of a marker rack. We specify this objective through a goal position $g_\text{marker-A} = \begin{bmatrix} 0.443, 0.014, 0.152 \end{bmatrix}^T$ for the end-effector. The reward function for this task is
\begin{align*}
    r^2(\bs, \ba) &= 1 - \tanh(10 \cdot \| \bs_{xyz} - g_\text{marker-A} \|_2).
\end{align*}

\noindent \textbf{Task 3: Eraser insertion.} The objective is to align the eraser to the corresponding hole of the same rack from the previous task. We specify this objective through a goal position $g_\text{eraser} = \begin{bmatrix} 0.448, 0.067, 0.152 \end{bmatrix}^T$ for the end-effector. The reward function for this task is
\begin{align*}
    r^3(\bs, \ba) &= 1 - \tanh(10 \cdot \| \bs_{xyz} - g_\text{eraser} \|_2),
\end{align*}

\noindent \textbf{Task 4: Bottle capping (A).} The objective is to align the cap to a Gatorade bottle. We specify this objective through a goal position $g_\text{bottle-A} = \begin{bmatrix} 0.469, 0.053, 0.189 \end{bmatrix}^T$ for the end-effector. Different from the previous reward functions, we additionally specify a waypoint $w^1 = \begin{bmatrix} 0.469, 0.053, 0.230 \end{bmatrix}^T$, as the bottle is significantly taller than the other objects from previous tasks. The reward function for this task is
\begin{align*}
    r^4(\st, \at) &= \frac{1}{2}(\mathbbm{1}(\| \bs_{xy} - w^1_{xy} \|_2 < 0.03) \cdot \mathbbm{1}(|\bs_z - w^1_z| \le 0.005) \\
    &+ (1 - \tanh(10 \cdot \| \bs_{xyz} - g_\text{bottle-A} \|_2))),
\end{align*}
where $\bs_{xy}$ are the $xy$-coordinates of the end-effector, $w^1_{xy}$ are the $xy$-coordinates of the waypoint, $\bs_z$ is the $z$-coordinate of the effector, and $w^1_z$ is the $z$-coordinate of the waypoint.

\noindent \textbf{Task 5: Bottle capping (B).} The objective is to align the cap to a plastic water bottle. We specify this objective through a goal position $g_\text{bottle-B} = \begin{bmatrix} 0.469, -0.014,  0.210 \end{bmatrix}$ for the end-effector. Similar to the previous bottle task, because this bottle is also relatively tall, we additionally specify a waypoint $w^2 = \begin{bmatrix} 0.469, -0.014,  0.240 \end{bmatrix}$. The reward function for this task is:
\begin{align*}
    r^5(\st, \at) &= \frac{1}{2}(\mathbbm{1}(\| \bs_{xy} - w^2_{xy} \|_2 < 0.03) \cdot \mathbbm{1}(|\bs_z - w^2_z| \le 0.005) \\
    &+ (1 - \tanh(10 \cdot \| \bs_{xyz} - g_\text{bottle-B} \|_2))),
\end{align*}
where $w^2_{xy}$ are the $xy$-coordinates of the waypoint and $w^2_z$ is the $z$-coordinate of the waypoint.

\noindent \textbf{Task 6: Block insertion (A).} The objective is to align the square block to the corresponding hole of the toy cube. We specify this objective through a goal position $g_\text{block-A} = \begin{bmatrix} 0.472, -0.002,  0.125 \end{bmatrix}^T$ for the end-effector. The reward function for this task is
\begin{align*}
    r^6(\bs, \ba) &= 1 - \tanh(10 \cdot \| \bs_{xyz} - g_\text{block-A} \|_2),
\end{align*}

\noindent \textbf{Task 7: Block insertion (B).} The objective is to align the parallelogram-shaped block to the corresponding hole of the toy cube. We specify this objective through a goal position $g_\text{block-B} = \begin{bmatrix} 0.465, -0.015,  0.140 \end{bmatrix}^T$ for the end-effector. The reward function for this task is
\begin{align*}
    r^7(\bs, \ba) &= 1 - \tanh(10 \cdot \| \bs_{xyz} - g_\text{block-B} \|_2),
\end{align*}

\noindent \textbf{Task 8: Block insertion (C).} The objective is to align the octagon-shaped block to the corresponding hole of the toy cube. We specify this objective through a goal position $g_\text{block-C} = \begin{bmatrix} 0.472, -0.060,  0.140 \end{bmatrix}^T$ for the end-effector. The reward function for this task is
\begin{align*}
    r^8(\bs, \ba) &= 1 - \tanh(10 \cdot \| \bs_{xyz} - g_\text{block-C} \|_2),
\end{align*}

\noindent \textbf{Task 9: Bottle capping (C).} The objective is to align the cap to the Vitamin water bottle. We specify this objective through a goal position $g_\text{bottle-C} = \begin{bmatrix} 0.460, -0.032,  0.185 \end{bmatrix}^T$ for the end-effector. The reward function for this task is
\begin{align*}
    r^9(\bs, \ba) &= 1 - \tanh(10 \cdot \| \bs_{xyz} - g_\text{bottle-C} \|_2),
\end{align*}

\noindent \textbf{Task 10: Marker insertion (B).} The objective is to align the cap to the Vitamin water bottle. We specify this objective through a goal position $g_\text{marker-B} = \begin{bmatrix} 0.444, -0.020,  0.118 \end{bmatrix}^T$ for the end-effector. The reward function for this task is
\begin{align*}
    r^9(\bs, \ba) &= 1 - \tanh(10 \cdot \| \bs_{xyz} - g_\text{marker-B} \|_2),
\end{align*}

All of the goal positions lie within a bounded region roughly of  size $2 \text{cm} \times 4 \text{cm} \times 4 \text{cm}$.

\section{Additional Plots}
\label{app:results}

\begin{figure*}[h]
    \centering
    \begin{minipage}{\linewidth}
    \centering
    \subfloat[Task 2: Marker insertion (A).]{
        \includegraphics[width=0.3\linewidth]{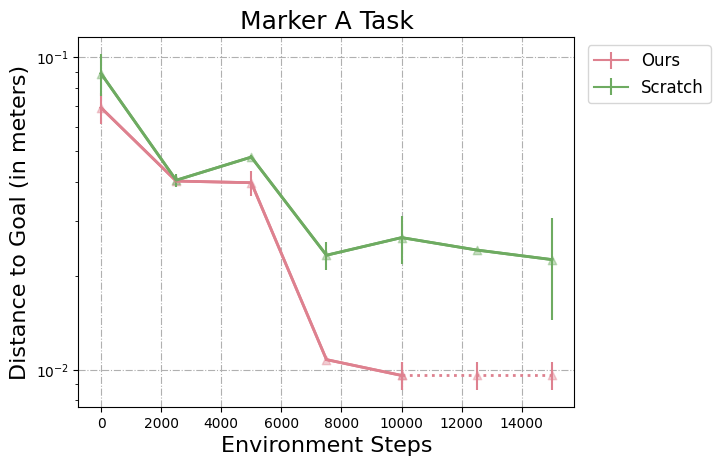}
        \label{fig:marker_a}
    } 
    \subfloat[Task 3: Eraser insertion.]{
        \includegraphics[width=0.3\linewidth]{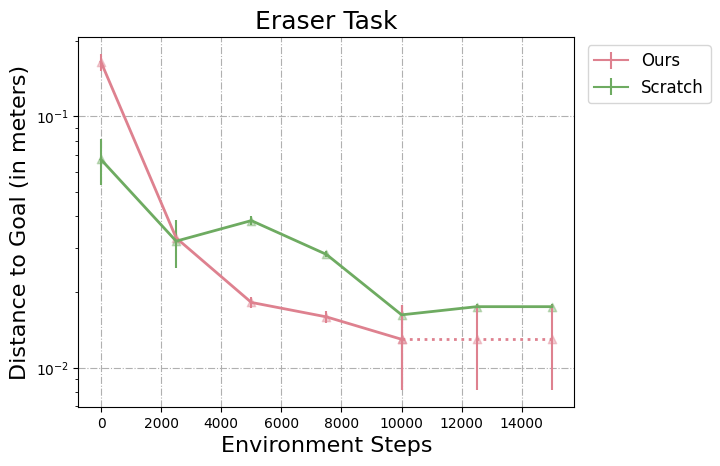}
        \label{fig:eraser}
    }
    \subfloat[Task 4: Bottle capping (A).]{
        \includegraphics[width=0.3\linewidth]{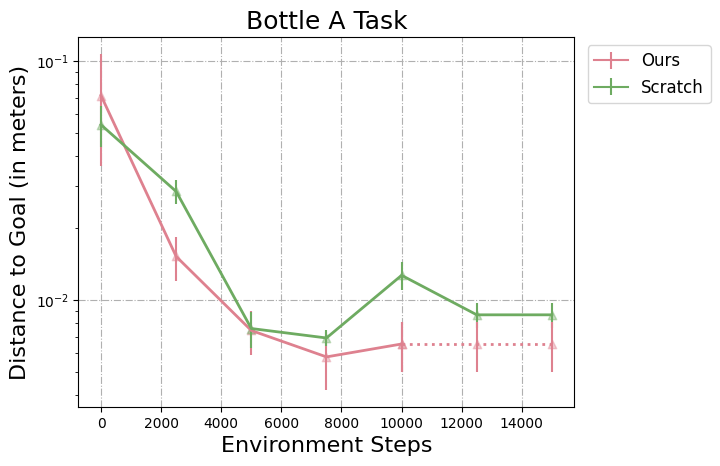}
        \label{fig:bottle_a}
    }\\
    \subfloat[Task 5: Bottle capping (B).]{
        \includegraphics[width=0.3\linewidth]{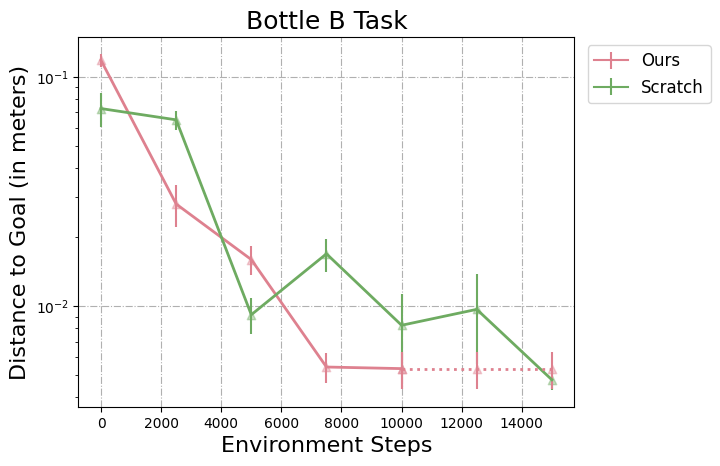}
        \label{fig:bottle_b}
    }
    \subfloat[Task 6: Block insertion (A).]{
        \includegraphics[width=0.3\linewidth]{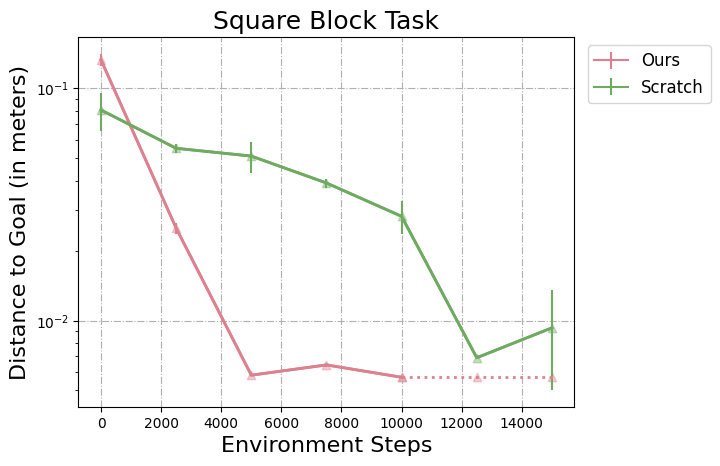}
        \label{fig:block_a}
    }
    \subfloat[Task 7: Block insertion (B).]{
        \includegraphics[width=0.3\linewidth]{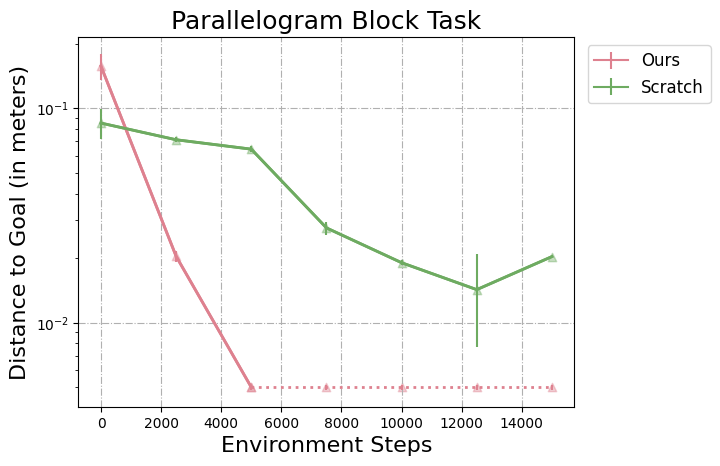}
        \label{fig:block_b}
    }\\
    \subfloat[Task 8: Block insertion (C).]{
        \includegraphics[width=0.3\linewidth]{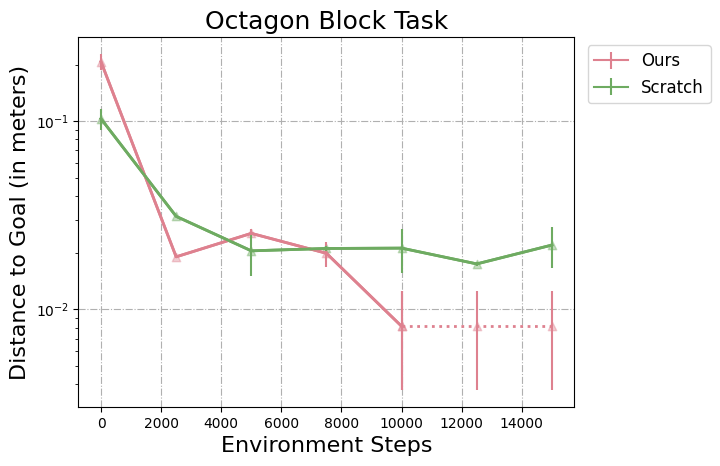}
        \label{fig:block_c}
    }
    \subfloat[Task 9: Bottle capping (C).]{
        \includegraphics[width=0.3\linewidth]{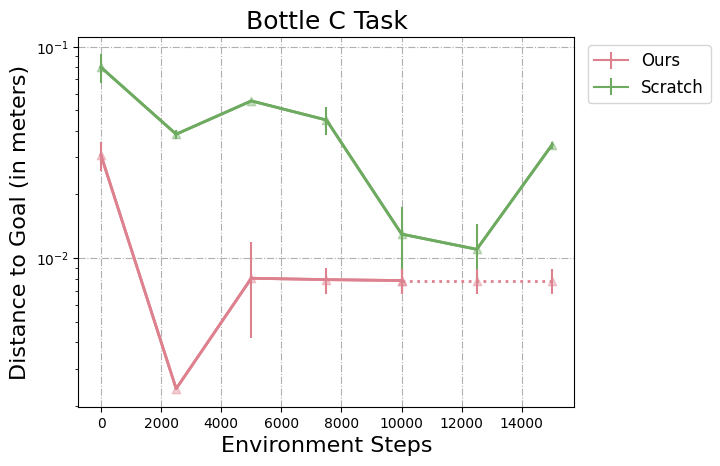}
        \label{fig:bottle_c}
    }
    \subfloat[Task 10: Marker insertion (B).]{
        \includegraphics[width=0.3\linewidth]{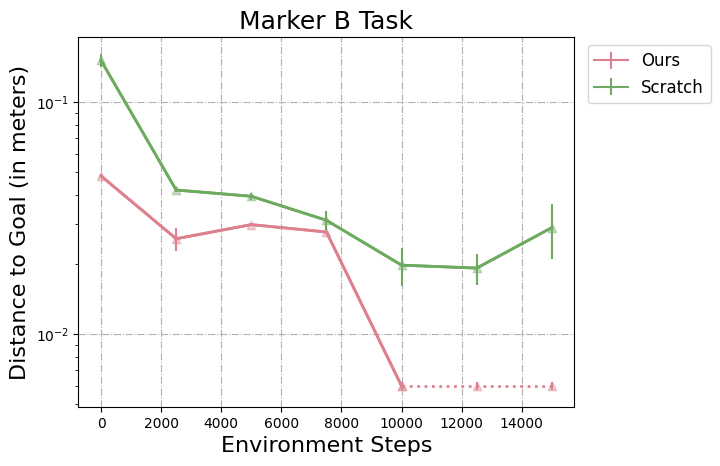}
        \label{fig:marker_b}
    }
    \caption{\small Individual learning curves for each task.}
    \label{fig:task_plots}
    \end{minipage}
\end{figure*}

In Fig.~\ref{fig:task_plots}, we provide the individual learning curves for each of the $10$ tasks from our robot experiments. For each data-point, we average the distance to the goal position (in meters) at the final time-step across $10$ evaluation episodes. Our method attains a lower average distance after $10$K time-steps than learning from scratch for all $5$ tasks following the initial reaching task.

%% file: main.bbl
\begin{thebibliography}{68}
\providecommand{\natexlab}[1]{#1}
\providecommand{\url}[1]{\texttt{#1}}
\expandafter\ifx\csname urlstyle\endcsname\relax
  \providecommand{\doi}[1]{doi: #1}\else
  \providecommand{\doi}{doi: \begingroup \urlstyle{rm}\Url}\fi

\bibitem[Aljundi et~al.(2019)Aljundi, Lin, Goujaud, and
  Bengio]{aljundi2019gradient}
Rahaf Aljundi, Min Lin, Baptiste Goujaud, and Yoshua Bengio.
\newblock Gradient based sample selection for online continual learning.
\newblock \emph{Advances in neural information processing systems}, 32, 2019.

\bibitem[Andrychowicz et~al.(2017)Andrychowicz, Wolski, Ray, Schneider, Fong,
  Welinder, McGrew, Tobin, Abbeel, and Zaremba]{andrychowicz2017hindsight}
Marcin Andrychowicz, Filip Wolski, Alex Ray, Jonas Schneider, Rachel Fong,
  Peter Welinder, Bob McGrew, Josh Tobin, Pieter Abbeel, and Wojciech Zaremba.
\newblock Hindsight experience replay.
\newblock \emph{Neural Information Processing Systems (NeurIPS)}, 2017.

\bibitem[Andrychowicz et~al.(2020)Andrychowicz, Baker, Chociej, Jozefowicz,
  McGrew, Pachocki, Petron, Plappert, Powell, Ray,
  et~al.]{andrychowicz2020learning}
OpenAI:~Marcin Andrychowicz, Bowen Baker, Maciek Chociej, Rafal Jozefowicz, Bob
  McGrew, Jakub Pachocki, Arthur Petron, Matthias Plappert, Glenn Powell, Alex
  Ray, et~al.
\newblock Learning dexterous in-hand manipulation.
\newblock \emph{The International Journal of Robotics Research}, 2020.

\bibitem[Arndt et~al.(2020)Arndt, Hazara, Ghadirzadeh, and
  Kyrki]{arndt2020meta}
Karol Arndt, Murtaza Hazara, Ali Ghadirzadeh, and Ville Kyrki.
\newblock Meta reinforcement learning for sim-to-real domain adaptation.
\newblock \emph{{IEEE} International Conference on Robotics and Automation
  (ICRA)}, 2020.

\bibitem[Baktashmotlagh et~al.(2014)Baktashmotlagh, Harandi, Lovell, and
  Salzmann]{baktashmotlagh2014domain}
Mahsa Baktashmotlagh, Mehrtash~T Harandi, Brian~C Lovell, and Mathieu Salzmann.
\newblock Domain adaptation on the statistical manifold.
\newblock \emph{{IEEE} Conference on Computer Vision and Pattern Recognition
  (CVPR)}, 2014.

\bibitem[Balaji et~al.(2020)Balaji, Farajtabar, Yin, Mott, and
  Li]{balaji2020effectiveness}
Yogesh Balaji, Mehrdad Farajtabar, Dong Yin, Alex Mott, and Ang Li.
\newblock The effectiveness of memory replay in large scale continual learning.
\newblock \emph{arXiv preprint arXiv:2010.02418}, 2020.

\bibitem[Bang et~al.(2021)Bang, Kim, Yoo, Ha, and Choi]{bang2021rainbow}
Jihwan Bang, Heesu Kim, YoungJoon Yoo, Jung-Woo Ha, and Jonghyun Choi.
\newblock Rainbow memory: Continual learning with a memory of diverse samples.
\newblock In \emph{Proceedings of the IEEE/CVF Conference on Computer Vision
  and Pattern Recognition}, pp.\  8218--8227, 2021.

\bibitem[Bickel et~al.(2007)Bickel, Br{\"u}ckner, and
  Scheffer]{bickel2007discriminative}
Steffen Bickel, Michael Br{\"u}ckner, and Tobias Scheffer.
\newblock Discriminative learning for differing training and test
  distributions.
\newblock In \emph{Proceedings of the 24th international conference on Machine
  learning}, pp.\  81--88, 2007.

\bibitem[Buzzega et~al.(2020)Buzzega, Boschini, Porrello, Abati, and
  Calderara]{buzzega2020dark}
Pietro Buzzega, Matteo Boschini, Angelo Porrello, Davide Abati, and Simone
  Calderara.
\newblock Dark experience for general continual learning: a strong, simple
  baseline.
\newblock \emph{Advances in neural information processing systems},
  33:\penalty0 15920--15930, 2020.

\bibitem[Caruana(1997)]{caruana1997multitask}
Rich Caruana.
\newblock Multitask learning.
\newblock \emph{Machine learning}, 28\penalty0 (1):\penalty0 41--75, 1997.

\bibitem[Chaudhry et~al.(2019)Chaudhry, Rohrbach, Elhoseiny, Ajanthan, Dokania,
  Torr, and Ranzato]{chaudhry2019tiny}
Arslan Chaudhry, Marcus Rohrbach, Mohamed Elhoseiny, Thalaiyasingam Ajanthan,
  Puneet~K Dokania, Philip~HS Torr, and Marc'Aurelio Ranzato.
\newblock On tiny episodic memories in continual learning.
\newblock \emph{arXiv preprint arXiv:1902.10486}, 2019.

\bibitem[Cortes et~al.(2008)Cortes, Mohri, Riley, and
  Rostamizadeh]{cortes2008sample}
Corinna Cortes, Mehryar Mohri, Michael Riley, and Afshin Rostamizadeh.
\newblock Sample selection bias correction theory.
\newblock In \emph{International conference on algorithmic learning theory},
  pp.\  38--53. Springer, 2008.

\bibitem[Devin et~al.(2017)Devin, Gupta, Darrell, Abbeel, and
  Levine]{devin2017learning}
Coline Devin, Abhishek Gupta, Trevor Darrell, Pieter Abbeel, and Sergey Levine.
\newblock Learning modular neural network policies for multi-task and
  multi-robot transfer.
\newblock \emph{{IEEE} International Conference on Robotics and Automation
  (ICRA)}, 2017.

\bibitem[Eysenbach et~al.(2021)Eysenbach, Asawa, Chaudhari, Salakhutdinov, and
  Levine]{eysenbach2020off}
Benjamin Eysenbach, Swapnil Asawa, Shreyas Chaudhari, Ruslan Salakhutdinov, and
  Sergey Levine.
\newblock Off-dynamics reinforcement learning: Training for transfer with
  domain classifiers.
\newblock \emph{International Conference on Learning Representations (ICLR)},
  2021.

\bibitem[Fern{\'a}ndez \& Veloso(2006)Fern{\'a}ndez and
  Veloso]{fernandez2006probabilistic}
Fernando Fern{\'a}ndez and Manuela Veloso.
\newblock Probabilistic policy reuse in a reinforcement learning agent.
\newblock In \emph{AAMAS}, pp.\  720--727, 2006.

\bibitem[Finn \& Levine(2017)Finn and Levine]{finn2017deep}
Chelsea Finn and Sergey Levine.
\newblock Deep visual foresight for planning robot motion.
\newblock In \emph{2017 IEEE International Conference on Robotics and
  Automation (ICRA)}, pp.\  2786--2793. IEEE, 2017.

\bibitem[Finn et~al.(2016)Finn, Levine, and Abbeel]{finn2016guided}
Chelsea Finn, Sergey Levine, and Pieter Abbeel.
\newblock Guided cost learning: Deep inverse optimal control via policy
  optimization.
\newblock \emph{International Conference on Machine Learning (ICML)}, 2016.

\bibitem[Finn et~al.(2017)Finn, Abbeel, and Levine]{finn2017model}
Chelsea Finn, Pieter Abbeel, and Sergey Levine.
\newblock Model-agnostic meta-learning for fast adaptation of deep networks.
\newblock In \emph{International Conference on Machine Learning}, 2017.

\bibitem[Fu et~al.(2018{\natexlab{a}})Fu, Luo, and Levine]{fu2017learning}
Justin Fu, Katie Luo, and Sergey Levine.
\newblock Learning robust rewards with adversarial inverse reinforcement
  learning.
\newblock \emph{International Conference on Learning Representations (ICLR)},
  2018{\natexlab{a}}.

\bibitem[Fu et~al.(2018{\natexlab{b}})Fu, Singh, Ghosh, Yang, and
  Levine]{fu2018variational}
Justin Fu, Avi Singh, Dibya Ghosh, Larry Yang, and Sergey Levine.
\newblock Variational inverse control with events: A general framework for
  data-driven reward definition.
\newblock \emph{Neural Information Processing Systems (NeurIPS)},
  2018{\natexlab{b}}.

\bibitem[Gu et~al.(2017)Gu, Holly, Lillicrap, and Levine]{gu2017deep}
Shixiang Gu, Ethan Holly, Timothy Lillicrap, and Sergey Levine.
\newblock Deep reinforcement learning for robotic manipulation with
  asynchronous off-policy updates.
\newblock In \emph{international conference on robotics and automation (ICRA)},
  2017.

\bibitem[Haarnoja et~al.(2018{\natexlab{a}})Haarnoja, Ha, Zhou, Tan, Tucker,
  and Levine]{haarnoja2018learning}
Tuomas Haarnoja, Sehoon Ha, Aurick Zhou, Jie Tan, George Tucker, and Sergey
  Levine.
\newblock Learning to walk via deep reinforcement learning.
\newblock \emph{Robotics: Science and Systems (RSS)}, 2018{\natexlab{a}}.

\bibitem[Haarnoja et~al.(2018{\natexlab{b}})Haarnoja, Zhou, Abbeel, and
  Levine]{haarnoja2018soft}
Tuomas Haarnoja, Aurick Zhou, Pieter Abbeel, and Sergey Levine.
\newblock Soft actor-critic: Off-policy maximum entropy deep reinforcement
  learning with a stochastic actor.
\newblock \emph{International Conference on Machine Learning (ICML)},
  2018{\natexlab{b}}.

\bibitem[Hausman et~al.(2018)Hausman, Springenberg, Wang, Heess, and
  Riedmiller]{hausman2018learning}
Karol Hausman, Jost~Tobias Springenberg, Ziyu Wang, Nicolas Heess, and Martin
  Riedmiller.
\newblock Learning an embedding space for transferable robot skills.
\newblock \emph{International Conference on Learning Representations (ICLR)},
  2018.

\bibitem[Isele \& Cosgun(2018)Isele and Cosgun]{isele2018selective}
David Isele and Akansel Cosgun.
\newblock Selective experience replay for lifelong learning.
\newblock In \emph{Proceedings of the AAAI Conference on Artificial
  Intelligence}, volume~32, 2018.

\bibitem[Julian et~al.(2020)Julian, Swanson, Sukhatme, Levine, Finn, and
  Hausman]{julian2020efficient}
Ryan Julian, Benjamin Swanson, Gaurav~S Sukhatme, Sergey Levine, Chelsea Finn,
  and Karol Hausman.
\newblock Efficient adaptation for end-to-end vision-based robotic
  manipulation.
\newblock \emph{Conference on Robot Learning (CoRL)}, 2020.

\bibitem[Kaelbling(1993)]{kaelblinglearning}
Leslie~Pack Kaelbling.
\newblock Learning to achieve goals.
\newblock \emph{IJCAI}, 1993.

\bibitem[Kalashnikov et~al.(2018)Kalashnikov, Irpan, Pastor, Ibarz, Herzog,
  Jang, Quillen, Holly, Kalakrishnan, Vanhoucke, et~al.]{kalashnikov2018qt}
Dmitry Kalashnikov, Alex Irpan, Peter Pastor, Julian Ibarz, Alexander Herzog,
  Eric Jang, Deirdre Quillen, Ethan Holly, Mrinal Kalakrishnan, Vincent
  Vanhoucke, et~al.
\newblock Qt-opt: Scalable deep reinforcement learning for vision-based robotic
  manipulation.
\newblock \emph{Conference on Robot Learning (CoRL)}, 2018.

\bibitem[Kirkpatrick et~al.(2017)Kirkpatrick, Pascanu, Rabinowitz, Veness,
  Desjardins, Rusu, Milan, Quan, Ramalho, Grabska-Barwinska,
  et~al.]{kirkpatrick2017overcoming}
James Kirkpatrick, Razvan Pascanu, Neil Rabinowitz, Joel Veness, Guillaume
  Desjardins, Andrei~A Rusu, Kieran Milan, John Quan, Tiago Ramalho, Agnieszka
  Grabska-Barwinska, et~al.
\newblock Overcoming catastrophic forgetting in neural networks.
\newblock \emph{Proceedings of the national academy of sciences}, 2017.

\bibitem[Kober et~al.(2013)Kober, Bagnell, and Peters]{kober2013reinforcement}
Jens Kober, J~Andrew Bagnell, and Jan Peters.
\newblock Reinforcement learning in robotics: A survey.
\newblock \emph{The International Journal of Robotics Research}, 2013.

\bibitem[Kohl \& Stone(2004)Kohl and Stone]{kohl2004policy}
Nate Kohl and Peter Stone.
\newblock Policy gradient reinforcement learning for fast quadrupedal
  locomotion.
\newblock In \emph{International Conference on Robotics and Automation}, 2004.

\bibitem[Krizhevsky et~al.(2012)Krizhevsky, Sutskever, and
  Hinton]{krizhevsky2012imagenet}
Alex Krizhevsky, Ilya Sutskever, and Geoffrey~E Hinton.
\newblock Imagenet classification with deep convolutional neural networks.
\newblock \emph{Advances in neural information processing systems}, 2012.

\bibitem[Lazaric(2012)]{lazaric2012transfer}
Alessandro Lazaric.
\newblock Transfer in reinforcement learning: a framework and a survey.
\newblock In \emph{Reinforcement Learning}, pp.\  143--173. Springer, 2012.

\bibitem[Lazaric et~al.(2008)Lazaric, Restelli, and
  Bonarini]{lazaric2008transfer}
Alessandro Lazaric, Marcello Restelli, and Andrea Bonarini.
\newblock Transfer of samples in batch reinforcement learning.
\newblock In \emph{international conference on Machine learning}, 2008.

\bibitem[Lee et~al.(2020)Lee, Hwangbo, Wellhausen, Koltun, and
  Hutter]{lee2020learning}
Joonho Lee, Jemin Hwangbo, Lorenz Wellhausen, Vladlen Koltun, and Marco Hutter.
\newblock Learning quadrupedal locomotion over challenging terrain.
\newblock \emph{Science robotics}, 5\penalty0 (47), 2020.

\bibitem[Lee et~al.(2019)Lee, Zhu, Srinivasan, Shah, Savarese, Fei-Fei, Garg,
  and Bohg]{lee2019making}
Michelle~A Lee, Yuke Zhu, Krishnan Srinivasan, Parth Shah, Silvio Savarese,
  Li~Fei-Fei, Animesh Garg, and Jeannette Bohg.
\newblock Making sense of vision and touch: Self-supervised learning of
  multimodal representations for contact-rich tasks.
\newblock In \emph{International Conference on Robotics and Automation (ICRA)}.
  IEEE, 2019.

\bibitem[Levine et~al.(2016)Levine, Finn, Darrell, and Abbeel]{levine2016end}
Sergey Levine, Chelsea Finn, Trevor Darrell, and Pieter Abbeel.
\newblock End-to-end training of deep visuomotor policies.
\newblock \emph{The Journal of Machine Learning Research}, 2016.

\bibitem[McCloskey \& Cohen(1989)McCloskey and
  Cohen]{mccloskey1989catastrophic}
Michael McCloskey and Neal~J Cohen.
\newblock Catastrophic interference in connectionist networks: The sequential
  learning problem.
\newblock In \emph{Psychology of learning and motivation}, volume~24, pp.\
  109--165. Elsevier, 1989.

\bibitem[Nagabandi et~al.(2019)Nagabandi, Clavera, Liu, Fearing, Abbeel,
  Levine, and Finn]{nagabandi2018learning}
Anusha Nagabandi, Ignasi Clavera, Simin Liu, Ronald~S Fearing, Pieter Abbeel,
  Sergey Levine, and Chelsea Finn.
\newblock Learning to adapt in dynamic, real-world environments through
  meta-reinforcement learning.
\newblock \emph{International Conference on Learning Representations (ICLR)},
  2019.

\bibitem[Parisotto et~al.(2016)Parisotto, Ba, and
  Salakhutdinov]{parisotto2016actor}
Emilio Parisotto, Lei~Jimmy Ba, and Ruslan Salakhutdinov.
\newblock Actor-mimic: Deep multitask and transfer reinforcement learning.
\newblock \emph{International Conference on Learning Representations (ICLR)},
  2016.

\bibitem[Riedmiller et~al.(2018)Riedmiller, Hafner, Lampe, Neunert, Degrave,
  Wiele, Mnih, Heess, and Springenberg]{riedmiller2018learning}
Martin Riedmiller, Roland Hafner, Thomas Lampe, Michael Neunert, Jonas Degrave,
  Tom Wiele, Vlad Mnih, Nicolas Heess, and Jost~Tobias Springenberg.
\newblock Learning by playing solving sparse reward tasks from scratch.
\newblock \emph{International Conference on Machine Learning (ICML)}, 2018.

\bibitem[Rolnick et~al.(2019)Rolnick, Ahuja, Schwarz, Lillicrap, and
  Wayne]{rolnick2018experience}
David Rolnick, Arun Ahuja, Jonathan Schwarz, Timothy~P Lillicrap, and Greg
  Wayne.
\newblock Experience replay for continual learning.
\newblock \emph{Neural Information Processing Systems (NeurIPS)}, 2019.

\bibitem[Rusu et~al.(2016{\natexlab{a}})Rusu, Colmenarejo, Gulcehre,
  Desjardins, Kirkpatrick, Pascanu, Mnih, Kavukcuoglu, and
  Hadsell]{rusu2015policy}
Andrei~A Rusu, Sergio~Gomez Colmenarejo, Caglar Gulcehre, Guillaume Desjardins,
  James Kirkpatrick, Razvan Pascanu, Volodymyr Mnih, Koray Kavukcuoglu, and
  Raia Hadsell.
\newblock Policy distillation.
\newblock \emph{International Conference on Learning Representations (ICLR)},
  2016{\natexlab{a}}.

\bibitem[Rusu et~al.(2016{\natexlab{b}})Rusu, Rabinowitz, Desjardins, Soyer,
  Kirkpatrick, Kavukcuoglu, Pascanu, and Hadsell]{rusu2016progressive}
Andrei~A Rusu, Neil~C Rabinowitz, Guillaume Desjardins, Hubert Soyer, James
  Kirkpatrick, Koray Kavukcuoglu, Razvan Pascanu, and Raia Hadsell.
\newblock Progressive neural networks.
\newblock \emph{arXiv preprint arXiv:1606.04671}, 2016{\natexlab{b}}.

\bibitem[Schmitt et~al.(2018)Schmitt, Hudson, Zidek, Osindero, Doersch,
  Czarnecki, Leibo, Kuttler, Zisserman, and Simonyan]{schmitt2018kickstarting}
Simon Schmitt, Jonathan~J Hudson, Augustin Zidek, Simon Osindero, Carl Doersch,
  Wojciech~M Czarnecki, Joel~Z Leibo, Heinrich Kuttler, Andrew Zisserman, and
  Karen Simonyan.
\newblock Kickstarting deep reinforcement learning.
\newblock \emph{arXiv preprint arXiv:1803.03835}, 2018.

\bibitem[Schwarz et~al.(2018)Schwarz, Czarnecki, Luketina, Grabska-Barwinska,
  Teh, Pascanu, and Hadsell]{schwarz2018progress}
Jonathan Schwarz, Wojciech Czarnecki, Jelena Luketina, Agnieszka
  Grabska-Barwinska, Yee~Whye Teh, Razvan Pascanu, and Raia Hadsell.
\newblock Progress \& compress: A scalable framework for continual learning.
\newblock \emph{International Conference on Machine Learning (ICML)}, 2018.

\bibitem[Singh et~al.(2019)Singh, Yang, Hartikainen, Finn, and
  Levine]{singh2019end}
Avi Singh, Larry Yang, Kristian Hartikainen, Chelsea Finn, and Sergey Levine.
\newblock End-to-end robotic reinforcement learning without reward engineering.
\newblock \emph{Robotics: Science and Systems (RSS)}, 2019.

\bibitem[Song et~al.(2020)Song, Yang, Choromanski, Caluwaerts, Gao, Finn, and
  Tan]{song2020rapidly}
Xingyou Song, Yuxiang Yang, Krzysztof Choromanski, Ken Caluwaerts, Wenbo Gao,
  Chelsea Finn, and Jie Tan.
\newblock Rapidly adaptable legged robots via evolutionary meta-learning.
\newblock \emph{{IEEE} International Conference on Intelligent Robots and
  Systems (IROS)}, 2020.

\bibitem[Sun et~al.(2021)Sun, Calandriello, Hu, Li, and
  Titsias]{sun2021information}
Shengyang Sun, Daniele Calandriello, Huiyi Hu, Ang Li, and Michalis Titsias.
\newblock Information-theoretic online memory selection for continual learning.
\newblock In \emph{International Conference on Learning Representations}, 2021.

\bibitem[Tao et~al.(2020)Tao, Genc, Sun, and Mallya]{tao2020repaint}
Yunzhe Tao, Sahika Genc, Tao Sun, and Sunil Mallya.
\newblock Repaint: Knowledge transfer in deep actor-critic reinforcement
  learning.
\newblock \emph{arXiv preprint arXiv:2011.11827}, 2020.

\bibitem[Taylor \& Stone(2009)Taylor and Stone]{taylor2009transfer}
Matthew~E Taylor and Peter Stone.
\newblock Transfer learning for reinforcement learning domains: A survey.
\newblock \emph{Journal of Machine Learning Research}, 10\penalty0 (7), 2009.

\bibitem[Taylor et~al.(2008)Taylor, Jong, and Stone]{taylor2008transferring}
Matthew~E Taylor, Nicholas~K Jong, and Peter Stone.
\newblock Transferring instances for model-based reinforcement learning.
\newblock In \emph{Joint European conference on machine learning and knowledge
  discovery in databases}, 2008.

\bibitem[Tedrake(2004)]{tedrake2004applied}
Russell~L Tedrake.
\newblock \emph{Applied optimal control for dynamically stable legged
  locomotion}.
\newblock PhD thesis, Massachusetts Institute of Technology, 2004.

\bibitem[Teh et~al.(2017)Teh, Bapst, Czarnecki, Quan, Kirkpatrick, Hadsell,
  Heess, and Pascanu]{teh2017distral}
Yee~Whye Teh, Victor Bapst, Wojciech~M Czarnecki, John Quan, James Kirkpatrick,
  Raia Hadsell, Nicolas Heess, and Razvan Pascanu.
\newblock Distral: Robust multitask reinforcement learning.
\newblock \emph{Neural Information Processing Systems (NeurIPS)}, 2017.

\bibitem[Tirinzoni et~al.(2018)Tirinzoni, Sessa, Pirotta, and
  Restelli]{tirinzoni2018importance}
Andrea Tirinzoni, Andrea Sessa, Matteo Pirotta, and Marcello Restelli.
\newblock Importance weighted transfer of samples in reinforcement learning.
\newblock In \emph{International Conference on Machine Learning}, 2018.

\bibitem[Todorov et~al.(2012)Todorov, Erez, and Tassa]{mujoco}
Emanuel Todorov, Tom Erez, and Yuval Tassa.
\newblock Mujoco: A physics engine for model-based control.
\newblock \emph{{IEEE} International Conference on Intelligent Robots and
  Systems (IROS)}, 2012.

\bibitem[Toussaint(2009)]{toussaint2009robot}
Marc Toussaint.
\newblock Robot trajectory optimization using approximate inference.
\newblock \emph{International Conference on Machine Learning (ICML)}, 2009.

\bibitem[Wang et~al.(2017)Wang, Kurth-Nelson, Tirumala, Soyer, Leibo, Munos,
  Blundell, Kumaran, and Botvinick]{wang2016learning}
Jane~X Wang, Zeb Kurth-Nelson, Dhruva Tirumala, Hubert Soyer, Joel~Z Leibo,
  Remi Munos, Charles Blundell, Dharshan Kumaran, and Matt Botvinick.
\newblock Learning to reinforcement learn.
\newblock \emph{CogSci}, 2017.

\bibitem[Wulfmeier et~al.(2015)Wulfmeier, Ondruska, and
  Posner]{wulfmeier2015maximum}
Markus Wulfmeier, Peter Ondruska, and Ingmar Posner.
\newblock Maximum entropy deep inverse reinforcement learning.
\newblock \emph{arXiv:1507.04888}, 2015.

\bibitem[Xie et~al.(2018)Xie, Singh, Levine, and Finn]{xie2018few}
Annie Xie, Avi Singh, Sergey Levine, and Chelsea Finn.
\newblock Few-shot goal inference for visuomotor learning and planning.
\newblock In \emph{Conference on Robot Learning}, 2018.

\bibitem[Yang et~al.(2022)Yang, Yang, Liu, and Sun]{yang2022evaluations}
Fan Yang, Chao Yang, Huaping Liu, and Fuchun Sun.
\newblock Evaluations of the gap between supervised and reinforcement lifelong
  learning on robotic manipulation tasks.
\newblock In \emph{Conference on Robot Learning}, pp.\  547--556. PMLR, 2022.

\bibitem[Yang et~al.(2020)Yang, Xu, Wu, and Wang]{yang2020multi}
Ruihan Yang, Huazhe Xu, Yi~Wu, and Xiaolong Wang.
\newblock Multi-task reinforcement learning with soft modularization.
\newblock \emph{Neural Information Processing Systems (NeurIPS)}, 2020.

\bibitem[Yin \& Pan(2017)Yin and Pan]{yin2017knowledge}
Haiyan Yin and Sinno Pan.
\newblock Knowledge transfer for deep reinforcement learning with hierarchical
  experience replay.
\newblock In \emph{Proceedings of the AAAI Conference on Artificial
  Intelligence}, volume~31, 2017.

\bibitem[Yu et~al.(2020)Yu, Kumar, Gupta, Levine, Hausman, and
  Finn]{yu2020gradient}
Tianhe Yu, Saurabh Kumar, Abhishek Gupta, Sergey Levine, Karol Hausman, and
  Chelsea Finn.
\newblock Gradient surgery for multi-task learning.
\newblock \emph{Neural Information Processing Systems (NeurIPS)}, 2020.

\bibitem[Zadrozny(2004)]{zadrozny2004learning}
Bianca Zadrozny.
\newblock Learning and evaluating classifiers under sample selection bias.
\newblock \emph{International Conference on Machine Learning (ICML)}, 2004.

\bibitem[Zhao et~al.(2020)Zhao, Nagabandi, Rakelly, Finn, and
  Levine]{zhao2020meld}
Tony~Z Zhao, Anusha Nagabandi, Kate Rakelly, Chelsea Finn, and Sergey Levine.
\newblock Meld: Meta-reinforcement learning from images via latent state
  models.
\newblock \emph{Conference on Robot Learning (CoRL)}, 2020.

\bibitem[Zhu et~al.(2020)Zhu, Wong, Mandlekar, and
  Mart\'{i}n-Mart\'{i}n]{robosuite2020}
Yuke Zhu, Josiah Wong, Ajay Mandlekar, and Roberto Mart\'{i}n-Mart\'{i}n.
\newblock robosuite: A modular simulation framework and benchmark for robot
  learning.
\newblock In \emph{arXiv preprint arXiv:2009.12293}, 2020.

\bibitem[Ziebart et~al.(2008)Ziebart, Maas, Bagnell, and
  Dey]{ziebart2008maximum}
Brian~D Ziebart, Andrew~L Maas, J~Andrew Bagnell, and Anind~K Dey.
\newblock Maximum entropy inverse reinforcement learning.
\newblock \emph{AAAI}, 2008.

\end{thebibliography}
